\documentclass{vgtc}         % ,onecolumn  review (journal style)

\usepackage{mathptmx}
\usepackage[pdftex]{graphicx}
\usepackage{epstopdf}
\usepackage{times}
\usepackage[lined,algonl,boxed]{algorithm2e}
\usepackage{amssymb,amsmath}
\usepackage{multirow}
\usepackage{subfigure}
\usepackage{paralist}
\usepackage{tikz}
\usepackage{multirow}
\usepackage{color}
\usepackage{ulem}
\usepackage{setspace}
\usepackage{microtype,hyphenat,balance}
\usepackage{stfloats}
\usepackage{array}
\usepackage{url}
\usepackage{bm} 
\usepackage{wrapfig}
\newcommand{\rev}[1]{\textcolor{black}{#1}}
\newcommand{\mengchen}[1]{\textcolor{black}{#1}}
\newcommand{\doc}[1]{\textcolor{black}{#1}}

\newcommand{\shixia}[1]{\textcolor{black}{#1}}

\onlineid{1023}

\vgtccategory{Research}

\vgtcinsertpkg

\title{Analyzing the \rev{Noise} Robustness of Deep Neural Networks}
 \author{Mengchen Liu$^{*}$, Shixia Liu$^{*}$, Hang Su$^{\dag}$, Kelei Cao\thanks{
e-mail: \{\{liumc13,ckl17\}@mails, shixia@mail\}.tsinghua.edu.cn. S. Liu is the corresponding author.}
, Jun Zhu\thanks{e-mail: \{suhangss,dcszj\}@mail.tsinghua.edu.cn}
 }

\affiliation{%
  \scriptsize
  $^*$School of Software, TNList Lab, State Key Lab for Intell. Tech. Sys., Tsinghua University
  \\
  
  $^{\dag}$Dept. of Comp. Sci.Tech., TNList Lab, State Key Lab for Intell. Tech. Sys., CBICR Center, Tsinghua University
}
\teaser{
\centering
\vspace{-3mm}
\centering
\includegraphics[width=\linewidth]{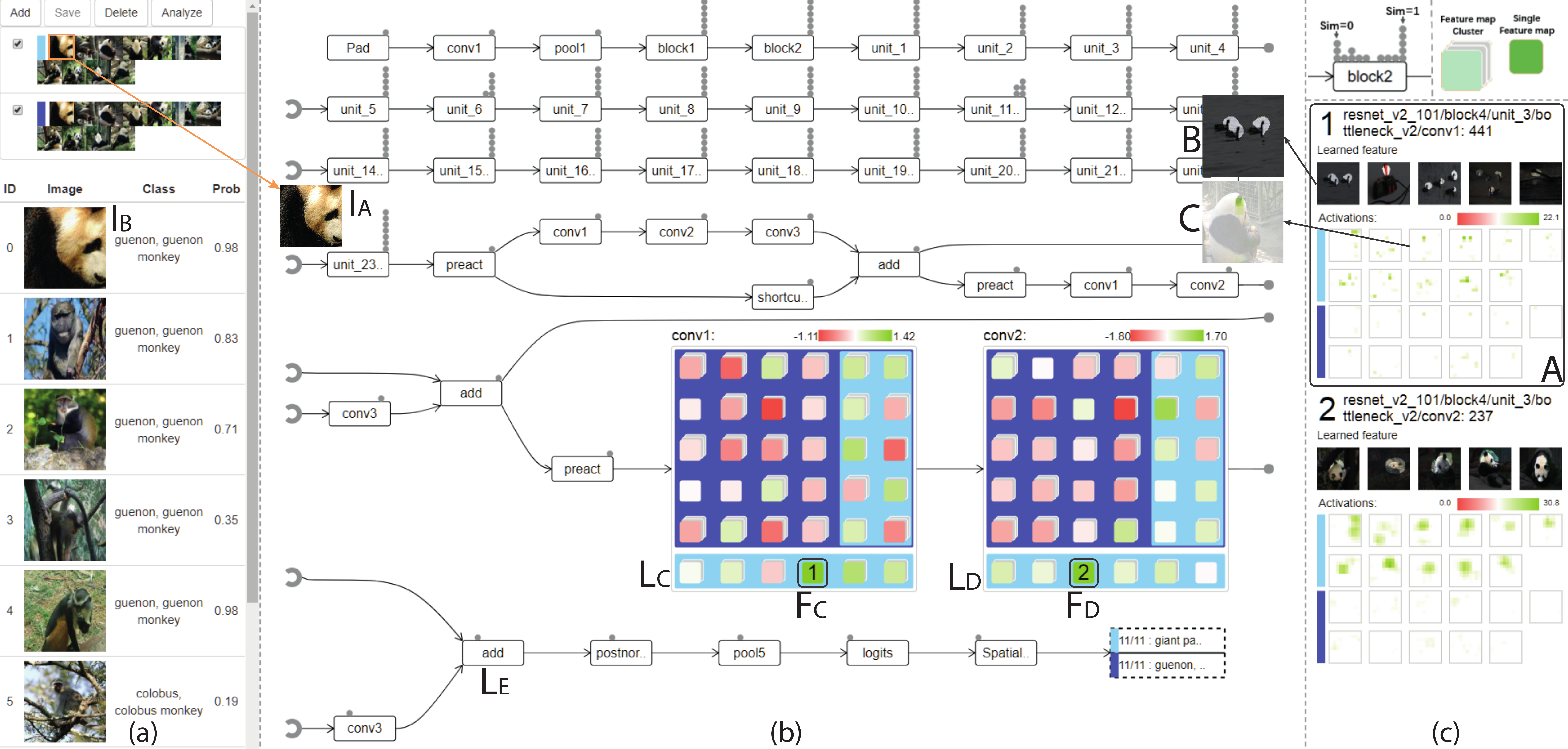}
\vspace{-4mm}
\caption{Explaining the misclassification of adversarial panda images. The root cause is that the CNN cannot \rev{detect} panda's ears in the adversarial examples ($F_C$), which leads to the failure of detecting \rev{a} panda's face ($F_D$). 
As a result, these adversarial examples are misclassified: (a) input images; (b) datapath visualization at the layer and feature map levels; (c) neuron visualization.
}
\label{fig:teaser}
}
\abstract{
Deep neural networks (DNNs) are vulnerable to maliciously generated adversarial examples.
\rev{These examples} are intentionally designed by making imperceptible perturbations and \rev{often} mislead a DNN \doc{into} making an \doc{incorrect} prediction.
This phenomenon means that there is significant risk in applying DNNs to safety-critical applications, such as driverless cars.
To address this issue, we present a visual analytics approach to \doc{explain} the primary cause of the wrong predictions introduced by adversarial examples.
\rev{The key is to analyze the datapaths of the adversarial examples and compare them with those of the normal examples.
A datapath is a group of critical neurons and their connections.}
To this end, we formulate the datapath extraction as a subset selection problem and approximately solve it based on back-propagation.
A multi-level visualization consisting of a segmented DAG (layer level), \rev{an Euler diagram} (feature map level), and a heat map (neuron level), \doc{has been} designed to help experts investigate datapaths from the high-level layers to the detailed neuron activations.
Two case studies are conducted \doc{that} demonstrate the promise of our approach in support of explaining the working mechanism of adversarial examples.\looseness=-1
}

\begin{document}
%\setlength{\parskip}{0.1pt}
%\setlength{\parskip}{3.1pt}

% \section{Related Work}
% \begin{itemize}
% 	\item Uncertainty Visualization
% 	\item Uncertainty Modeling
% 	\item Multivariate Visualization
% \end{itemize}

{
%\fontsize{8}{8} %this command adjusts the fontsize of all the mathmatical symbols in the paper
% !TEX root = EvoTree-KDD.tex

\firstsection{Introduction}
\maketitle

%\mengchen{Deep neural networks (DNNs) have demonstrated significant improvements over traditional approaches in many artificial intelligence tasks~\cite{Lecun2015_Survey}, such as image classification~\cite{He2016_CVPR_residual} and language translation~\cite{Sutskever2014_NIPS_Sequence}.}

Deep neural networks (DNNs) have evolved to become state-of-the-art in a torrent of artificial intelligence applications, such as image classification and language translation~\cite{Lecun2015_Survey,liu2017towards,wu2018ForVizor,xie2018semantic}.
%the 
However, researchers have recently found that DNNs are generally vulnerable to maliciously generated adversarial examples, which are intentionally designed to mislead a DNN \doc{into making incorrect} predictions~\cite{Dezfooli2017_CVPR_Universal, Nguyen2015_CVPR_Deep,szegedy2013intriguing,Zheng2016_CVPR}.
%\doc{into} make wrong 
For example, an attacker can modify an image of \doc{a} panda ($I_A$ in Fig.~\ref{fig:teaser}) slightly, even imperceptibly to human eyes, and
the generated adversarial example ($I_B$ in Fig.~\ref{fig:teaser}) is able to mislead a state-of-the-art DNN~\cite{He2016_CVPR_residual} to classify it as something else entirely (e.g., a monkey), \mengchen{because the DNN detects a monkey's face in the top right corner of the adversarial example (Fig.~\ref{fig:case0_monkey}A)}.
This phenomenon brings high risk in applying DNNs to safety- and security-critical applications, such as driverless cars, facial recognition ATMs, and Face ID security on mobile phones~\cite{Akhtar2018_threat}.
%To tackle this issue, there \doc{have been} some initial efforts in defending against adversarial examples~\cite{Cisse2017_ICML_parseval}.
%are 
%However, due to the limited understanding of the working mechanism of adversarial examples, the proposed defense strategies \doc{have} often \doc{been} very heavy, which usually leads to performance degradation.
%are
% \mengchen{For example, Dong et al. added a regularization to all the neurons in the layer that extracts the highest-level features.
% Although this defense approach can improve the robustness of a DNN, it causes a 4$\%$ drop in accuracy.
% %Finding the critical neurons in that layer and designing more targeted defense strategies will potentially decrease the performance degeneration.}
% Thus, there is a growing need to find the critical neurons and thus understand the primary cause of the wrong predictions introduced by adversarial examples~\cite{Akhtar2018_threat}, which \doc{can facilitate} experts to design more targeted and effective defense strategies.}
\shixia{Hence, there is a growing need to understand the inner workings of adversarial examples and identify the root cause of the incorrect predictions.}
%facilitates

% \mengchen{In this paper, we present a visual analytics approach to 
% As currently, almost all research on ADV is focused on CNNs,
% We take one CNN as an example}
There are two technical challenges to understanding and analyzing adversarial examples, \rev{which are derived from the discussions with machine learning experts (Sec.~\ref{sec:system}) and previous research on adversarial examples~\cite{dong2017boosting,Dezfooli2017_CVPR_Universal,Nguyen2015_CVPR_Deep}.}
%Ori: The first challenge is to extract the datapath for adversarial examples, which \doc{includes} the critical neurons and their connections that are responsible for the predictions of the examples (Fig.~\ref{fig:system} (b)).
\rev{The first challenge is how to extract the datapath for adversarial examples.
A datapath includes the critical neurons and their connections that are responsible for the predictions of the examples (Fig.~\ref{fig:system} (a)).}
%is
Disclosing the datapath will help experts design more targeted \doc{defense} approaches.
%defending
However, in a DNN, the neurons have complex interactions with each other~\cite{Bengio2013_PAMI_Representation}.
Thus, it is technically demanding to disentangle these neurons from the whole network and thus form the datapath.
% which neurons are responsible for the wrong prediction.
% Because there are often millions of neurons in a deep model and these neurons often influence each other, accurately and effectively finding such neurons are technically demanding.
% For brevity, we denote the set of such neurons as the \textbf{datapath} and the first challenge can be abbreviated as \textbf{datapath extraction}.
%The second challenge is effectively visualizing datapaths.
The second challenge is how to effectively illustrate the inner workings of adversarial examples based on the extracted datapaths.
%an effective exploration mechanism for experts to examine and compare the datapath
%With the development of more effective training approaches, CNNs are becoming more complex.
A state-of-the-art DNN usually contains hundreds of layers, with millions of neurons in each layer~\cite{He2016_CVPR_residual}.
%Ori: First, a CNN may consist of tens or hundreds of layers (depth), with thousands of neurons (width) in each layer, as well as millions of connections between neurons
Thus, an extracted datapath potentially contains millions of neurons \rev{and} even more connections.
%Moreover, a feature map usually contains thousands of neurons.
Directly visualizing all the neurons and the corresponding connections in a datapath will lead to excessive visual clutter.
%Ori:Directly visualizing all the time series with a line chart will induce excessive visual clutter

% multi-level
% many neurons

To tackle these challenges, we have developed a visual analytics tool, AEVis, to explain the root cause of the wrong predictions introduced by adversarial examples.
The key is to effectively extract and understand the datapath of adversarial examples.
We formulate the datapath extraction as a subset selection problem, which is NP-complete~\cite{cormen2009introduction}.
To analyze the adversarial examples in large DNNs, we approximate the subset selection problem as \doc{an} easy-to-solve quadratic optimization by Taylor decomposition\rev{~\cite{rudin1964principles}}, and solve the quadratic optimization using back-propagation~\cite{christopher2006_PRML}.
%a more 
Based on the extracted datapaths, we design a multi-level visualization that enables experts to effectively explore and compare datapaths from the high-level layers to the detailed neuron activations.
In particular, at the layer-level, we design a segmented directed acyclic graph (DAG) visualization to provide an overview of the datapaths \rev{(Fig.~\ref{fig:teaser} (b))}.
\rev{As shown in Fig.~\ref{fig:teaser} (c),} the detailed neuron activations are \rev{presented} as heat maps that are familiar to machine learning experts (neuron level).
%A dot plot is combined with the segmented DAG visualization, acting as a visual hint to enable experts select the layer of interest.
Between the layer level visualization and neuron level, we \doc{have added} a feature map level because a layer may contain millions of critical neurons.
%add 
%An expert can select multiple layers and analyze the critical feature maps of these layers in more detail.
\rev{A DNN, especially a convolutional neural network (CNN), organizes neurons in feature maps, each of which is a set of neurons sharing the same weight and thus detecting the same feature.
This inherent property enables the features to be recognized regardless of their position in the input (e.g., an image) and thus improves the generalization of DNNs~\cite{goodfellow2016deep}.}
At the feature map level, we employ \rev{an Euler diagram} to illustrate and compare critical feature maps belonging to different datapaths.
% The next sentence will be refined according to case study
Two case studies are conducted to demonstrate that our approach can better explain the working mechanism of both white-box and black-box adversarial examples.
% , and thus can facilitate experts in designing a more robust DNN
% \mengchen{We conduct two case studies to demonstrate how our approach helps experts determine whether an image is an adversarial example or not, explain the wrong predictions introduced by such adversarial examples, and thus design more robust CNNs.}

The key technical contributions of this \doc{paper} are:
%work
\begin{compactitem}
\item\noindent{\textbf{\normalsize A visual analytics tool} that explains the primary cause of the wrong predictions introduced by adversarial examples.}
\item\noindent{\textbf{\normalsize A datapath extraction method} that discloses critical \rev{neurons} and their connections that are responsible for a prediction.}
\item\noindent{\textbf{\normalsize A multi-level datapath visualization} that enables experts to examine and compare datapaths, from the high-level layers to the detailed neuron activations.}
%\item\noindent{\textbf{A credit assignment algorithm} that explains how the neurons contribute to a training failure.}
\end{compactitem}
 
In this paper, we focus on analyzing adversarial examples generated for CNNs \rev{on the task of image classification}, because currently most adversarial example generation approaches \mengchen{focus on} attacking CNNs \rev{on the image classification task}~\cite{Akhtar2018_threat} .
\rev{Besides CNNs, AEVis can be directly used to analyze other types of deep models, such as multilayer perceptrons (MLPs).}
% The techniques that focus on attacking deep learning in the real-world conditions 

% Challenge:

% 1. Extract datapath: 
% 1.1 determine the contribution of a feature map to a classification result
% 1.2 determine the influence of a feature map to another (to mention DGMTracker?)

% 2. Visualize the datapath and how images "pass" the datapath
% 2.1 scalability: 100+ layers, 1000+ feature maps in one layer
% 2.2 compare datapaths to illustrate the difference between normal examples and adversarial examples

% 1. datapath generation
% Step 1: feature maps (or neurons?) based on LIME (KDD 2016)
% Step 2: also based on LIME
% %connections based on Taylor decomposition (or partial derivatives)

% 2. visualization
% Step 1: DAG
% Step 2: segmentation based on DP
% Step 3: neurons based on Euler diagram
% Step 4: connections based on biclustering 

% Contribution:

% 1. datapath generation
% 2. focus+context datapath visualization, combining DAG, DAG segmentation (partition?), Euler diagram, biclustering
% !TEX root = EvoTree-KDD.tex

\section{Related Work}\label{sec:related-work}

\subsection{Visual Analytics for Explainable Deep Learning}
%cited 7 TVCG 2017-2018

\rev{A number of visual analytics approaches~\cite{bilal2018convolutional, liu2018deeptracker,liu2018analyzing, ming2017understanding, pezzotti2018deepeyes, rauber2017visualizing, strobelt2018lstmvis, wongsuphasawat2018visualizing} have been developed to illustrate the working mechanism of DNNs.
A comprehensive survey on exploring the space of interpretability interfaces was presented by Olah et al.~\cite{olah2018building}.}
%Ori: In the field of visual analytics, a number of approaches have been developed to illustrate the working mechanism of DNNs~\cite{bilal2018convolutional, liu2018analyzing, ming2017understanding, pezzotti2018deepeyes, rauber2017visualizing, strobelt2018lstmvis, wongsuphasawat2018visualizing}.
Most recent approaches \rev{on explainable deep learning} can be categorized into two groups: network-centric~\cite{liu2017towards,Tzeng2005_VIS,wongsuphasawat2018visualizing} and example-centric approaches~\cite{Harley2015_ISVC,kahng2018acti}.\looseness=-1

Network-centric approaches focus on illustrating the network structure of a DNN.
%Ori; Pioneering research was conducted by Tzeng et al.
%As one of the pioneering efforts in this direction, 
Tzeng et al. employed a DAG visualization to illustrate the neurons and their connections.
In particular, each neuron is represented by a node and their connections are represented by edges.
Their method can illustrate the structure of a small neural \doc{network}, but suffers from severe visual clutter when visualizing \doc{state-of-the-art} DNNs.
%well, the 
To solve this problem, Liu et al.~\cite{liu2017towards} developed a scalable visual analytics tool, CNNVis, based on clustering techniques. 
It helps machine learning experts to diagnose a failed training process.
%They provided a scalable visualization of deep CNNs based on clustering techniques.
Wongsuphasawat et al.~\cite{wongsuphasawat2018visualizing} developed a tool with a scalable graph visualization \rev{(TensorFlow Graph Visualizer)} to present the dataflow graph of a DNN.
To produce a legible graph visualization, they apply a set of graph transformations that converts the low-level dataflow graph to the high-level structure of a DNN.
The aforementioned approaches facilitate experts in better understanding the network structure, but they are less capable of explaining the predictions of individual examples.
\rev{For example, the TensorFlow Graph Visualizer developed by Wongsuphasawat et al. ~\cite{wongsuphasawat2018visualizing} does not extract and disclose the datapath of a set of examples, which is critical for identifying the root cause of the misclassification produced by adversarial examples.}
%which are necessary to find the root cause of a misclassification~\cite{kahng2018acti}

% Compared with these approaches, our work is able to explain the working mechanism of a prediction by extracting and illustrate the datapath of a set of examples.
% \mengchen{Such explanation can facilitate experts in finding the critical neurons for a wrong prediction and thus designing more robust models.}

There are several recent attempts to explain how DNNs make predictions for examples (example-centric approaches).
%Ori: that explains how models made predictions for each instance
A widely used approach is to feed a set of examples into a DNN, and visualize the internal activations produced by the examples.
For example, Hartley et al.~\cite{Harley2015_ISVC} developed an interactive node-link visualization to show the activations in a DNN.
%In their visualization, each node represents the activations on a feature map and \doc{they are presented} as a heat map.
%is shown
Although this method is able to illustrate detailed activations on feature maps, it suffers from severe visual clutter when dealing with large CNNs.
% Original:  Although their visualization is able to illustrate how data flows through a network, this method suffers from severe visual clutter when dealing with large networks
To solve this problem, Kahng et al.~\cite{kahng2018acti} developed ActiVis to interpret large-scale DNNs and their results.
% Ori: an interactive visualization system for interpreting large-scale deep learning models and results
They employed a multiple coordinated visualization to facilitate experts in comparing activations among examples and reveal the causes for misclassification. 
%, including an overview of the network structure, a neuron activation view, and a projected view showing examples distribution in a layer.
%The coordinated views combine 
Although ActiVis can show the causes for misclassification to some extent, it cannot be directly used to illustrate the primary causes of the wrong prediction caused by adversarial examples as it heavily relies on \doc{expert} knowledge to select \doc{which} layer to examine.
%experts'  a
In addition, we argue that purely relying on activations in one layer will result in misleading results (Sec.~\ref{sec:eval_dp}).
To solve this problem, we propose \doc{combining} activations and gradients for selecting critical neurons at different layers, which are connected to form a datapath for the adversarial examples of interest.
%to combine together
%\mengchen{which heavily depends on the expertise of the expert.}
% Original: This step heavily depends on the expertise of the expert.
In addition, we integrate a DAG visualization with dot plots, which provide guided visual hints to help experts select the layer of interest.\looseness = -1
%In addition, ACTIVIS employs activations to select critical neurons.

%Their work is More relevant to our work

% network centric: visualizing neurons, connections, ... e.g., cnnvis, 

% 2. we provide whole data path instead of the XXX in one / several layer (not sure)

% 3. focus on robustness

% 4. scalability

\subsection{Adversarial Attacks on Deep Learning}

%Recently, researchers find that DNNs, especially CNNs, are vulnerable to adversarial examples~\cite{szegedy2013intriguing}.
% The research work on adversarial attacks can be categorized into three groups~\cite{Akhtar2018_threat}: adversarial example generation~\cite{Dezfooli2017_CVPR_Universal, Nguyen2015_CVPR_Deep}, defense against adversarial examples~\cite{Cisse2017_ICML_parseval,Zheng2016_CVPR}, and theoretical analysis of adversarial examples~\cite{Goodfellow2015_ICLR_explaining, tanay2016boundary}.

Adversarial attacks are a new research focus of DNNs~\cite{Akhtar2018_threat}. 
Existing efforts mainly focus on the generation of adversarial examples~\cite{Dezfooli2017_CVPR_Universal, Nguyen2015_CVPR_Deep} and \doc{how to defend} against adversarial examples~\cite{Cisse2017_ICML_parseval,Zheng2016_CVPR}.
%is defense
%mengchen: no problem

%The works on generating adversarial examples acts as the basis of this field.
%Given a normal example, 
The key of generating an adversarial example is to find a very small perturbation that can mislead DNNs into misclassification.
Recently, researchers have proposed a variety of approaches \doc{for} finding such perturbations, including the Fast Gradient Sign Method~\cite{Goodfellow2015_ICLR_explaining}, universal adversarial perturbation~\cite{Dezfooli2017_CVPR_Universal}, and DeepFool~\cite{Moosavi-Dezfooli2016_CVPR_Simple}.
%to 
% For example, Moosavi-Dezfooli et al.~\cite{Moosavi-Dezfooli2016_CVPR_Simple} designed the DeepFool algorithm.
% The algorithm is an iterative process, where in each iteration the input example moves a very small step to the nearest decision boundary of the target DNN.
% The iterative process stops until the prediction of the example is changed.
% Our work takes the generated adversarial examples as input and aims at analyzing the working mechanisms of the wrong predictions introduced by adversarial examples.
% , which helps design more robust DNNs.

% Since the discovery of adversarial examples for DNNs, researchers have proposed various methods to defend against adversarial exmaples.
The generation of adversarial examples has \rev{inspired} researchers to develop several methods to defend against adversarial attacks.
%triggers
%Based on the idea that \mengchen{a machine learning algorithm cannot learn what it has not seen, }
%Mengchen: the above statement will be justified!! I will find a reference
A natural defense is training a DNN \doc{using} adversarial examples (adversarial training)~\cite{Goodfellow2015_ICLR_explaining, szegedy2013intriguing}.
%approach by 
Although adversarial training can defend the training adversarial examples, Moosavi-Dezfooli et al.~\cite{Dezfooli2017_CVPR_Universal} discovered that new types of adversarial examples can be generated to attack DNNs \doc{that have been trained in this way}.
%adversarially trained 
To tackle this issue, a more effective strategy is \doc{to change} the network structure to improve the defense against unseen adversarial examples.
%changing 
%Moosavi-Dezfooli [16] showed that effective adversarial examples can again be computed for already adversarially trained networks.
\doc{Adding} regularization to the corresponding layer(s) is one of the most commonly used \doc{methods for this}.
%Among these defense methods,  
The typical regularization methods include input gradient regularization~\cite{ross2017improving} and layer-wise Lipschitz regularization~\cite{Cisse2017_ICML_parseval}.
%drucker1992improving,
%and highest-level feature regularization~\cite{dong2017towards}. 
% Camera remove for space
%For example, Cisse et al.~\cite{Cisse2017_ICML_parseval} added a layer-wise regularization to control to the global Lipschitz constant~\cite{Lipschitz_continuity} of a DNN, and thus increased the robustness of a DNN to \doc{defend against} small perturbations in the input space.
% Ori: a layerwise regularization method for reducing the network’s sensitivity to small perturbations by carefully controlling its global Lipschitz constant
However, due to the limited understanding of the working mechanism of adversarial examples, the above defense strategies are often very heavy, which usually leads to performance degradation for large, complex datasets such as ImageNet~\cite{russakovsky2015imagenet}.
To solve this problem, we \doc{have developed} a visual analytics tool to explain the primary cause of the wrong predictions introduced by adversarial examples. 
%which facilitates experts in designing more targeted and effective defense strategies based on a few neurons.
Based on a few vulnerable neurons identified by AEVis (Fig.~\ref{fig:teaser}A), machine learning experts can design more targeted and effective defense strategies.

\section{The Design of AEVis}\label{sec:system}

\begin{figure*}[!t]
  \centering
  \vspace{-2mm}
  \includegraphics[width=0.93\linewidth]{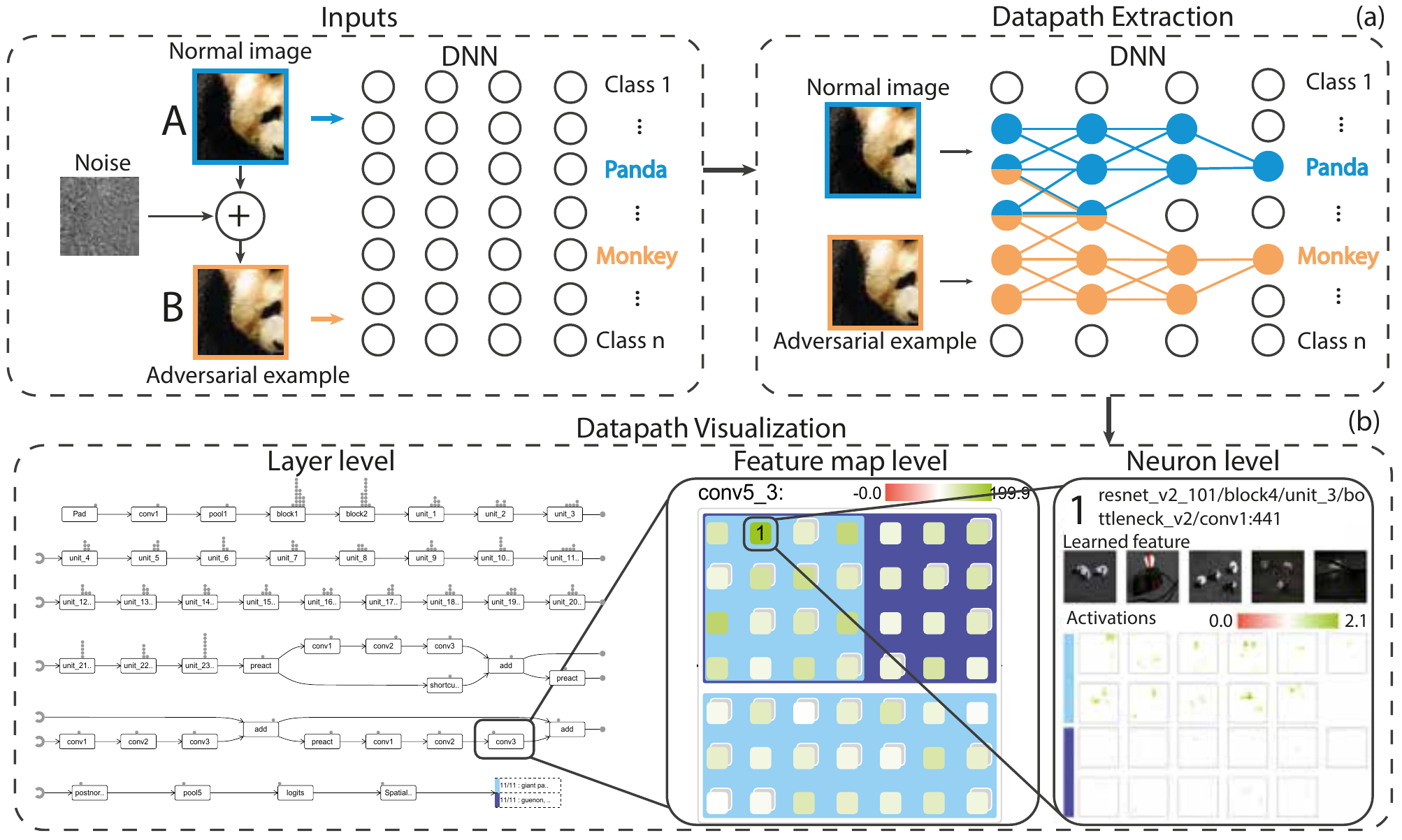}
  \vspace{-2mm}
  \caption{AEVis contains two modules: \rev{(a)} a datapath extraction module and \rev{(b)} a datapath visualization module that illustrates datapaths in multiple levels: layer level, feature map level, and neuron level.}\looseness=-1
  \label{fig:system}
  \vspace{-4mm}
\end{figure*}

\subsection{Motivation}
% Mengchen commetn: mention CONNECTIONS???? 
The development of AEVis is collaborated with the machine learning team that won the first place in the NIPS 2017 non-targeted adversarial attack and targeted adversarial attack competitions, which aim at attacking CNNs~\cite{dong2017boosting,NIPS2017_Challenge}.
%Despite their promising results, the experts complained to us about the inefficiency and inconvenience of their research process.
Despite their promising results, the experts found that the research process was inefficient and inconvenient, especially the explanation of the model outputs.
In their research process, a central step is explaining misclassification introduced by adversarial examples.
%As ``A fault confessed is half redressed''~\cite{dong2017towards},
Understanding why an error has been made help\rev{s} the experts detect the weakness of the model and further design a more effective attacking/defending approach.
To this end, they desire to understand the roles of the neurons and their connections for prediction.
Because there are millions of neurons in a CNN, examining all neurons and their connections is prohibitive.
In the prediction of a set of examples, the experts usually \textbf{extract} and \textbf{examine} the critical neurons and their connections, which are referred to as \textbf{datapaths} in their field.

To extract datapaths, the experts often treat the most activated neurons as the critical neurons~\cite{Zeiler2014_ECCV}.
However, they are not satisfied with the current activation-based approach because it may result in misleading results.
For instance, considering an image with highly recognizable secondary objects, which are mixed with the main object in the image.
% For instance, considering an image with an main object in the center and several surrounding objects, 
%there are sometimes one central object and multiple surrounding objects in the images of the ImageNet ILSVRC 2012 dataset~\cite{russakovsky2015imagenet}.
%There are high activations on the neurons that detect the surrounding objects, but the experts are not interested in them because these neurons are often irrelevant to the prediction of the central object.
The activations of the neurons that detect the secondary objects are also large, however, the experts are not interested in them because these neurons are often irrelevant to the prediction of the main object.
Currently, the experts have to rely on their knowledge to manually ignore \rev{these} neurons in the analysis process.\looseness=-1
%the

After extracting datapaths, the experts examine them to understand their roles for prediction.
%Using the existing 
%Currently, they are able to understand the roles of individual neuron and several connections well, using the LC metric~\cite{dong2017towards}, discrepancy maps~\cite{zhou2014object}, heat maps~\cite{Zeiler2014_ECCV}, weight visualization~\cite{Goodfellow2015_ICLR_explaining}.
Currently, they utilize discrepancy maps~\cite{zhou2014object}, heat maps~\cite{Zeiler2014_ECCV}, and weight visualization~\cite{Goodfellow2015_ICLR_explaining} to understand the role of the datapaths.
%the LC metric~\cite{dong2017towards}, 
Although these methods can help the experts at the neuron level, they commented that there lacked an effective exploration mechanism enabling them to investigate the extracted datapaths from high-level layers to individual neurons.

\subsection{Requirement Analysis}
\label{sec:require}

\rev{To collect the requirements of our tool, we follow the human-centered design process~\cite{brehmer2014overview,lloyd2011human}, which involves two experts (E$_1$ and E$_2$) from the winning team of the NIPS 2017 competition.
The design process consists of several iterations.
In each iteration, we present the developed prototype to the experts, probe further requirements, and modify our tool accordingly.
We have identified the following high-level requirements in this process.}
\rev{Among these requirements, \textbf{\normalsize{R2}} and \textbf{\normalsize{R3}} are two initial requirements, while \textbf{\normalsize{R1}} and \textbf{\normalsize{R4}} are gradually identified in the development.}  
% Ori: We have identified the following high-level requirements based on previous research and the discussions with two experts (E$_1$ and E$_2$) from the winning team of the NIPS 2017 competition.

\noindent \textbf{\normalsize R1 - Extracting the datapath for a set of examples of interest.}
Both experts expressed the need for extracting the datapath of an example, which serves as the basis for analyzing why an adversarial example is misclassified.
In a CNN, different neurons learn to detect different features~\cite{Zeiler2014_ECCV}.
Thus, the roles of the neurons are different for the prediction of an example.
E$_1$ said \rev{that analyzing} the datapath can greatly save experts' effort because they are able to focus on the critical neurons instead of examining all neurons.
% Mengchen comment: another reason is revealing the pattern, if displaying all neurons, the pattern can be masked
%Disclosing these responsible feature maps and their connections can largely reduce the effort needed because experts can focus on these critical feature map instead of tediously examining all feature maps.
Besides the datapath for individual examples, E$_1$ emphasized the need for extracting the common datapath for a set of examples of the same class.
He commented that the datapath of one example sometimes is not representative for the image class.
% For example, given a side photo of an panda, the extracted datapath will probably not including the neuron detecting the face of a panda, which is an very important feature to classify a panda.
For example, given an image of a panda's face, the extracted datapath will probably not \rev{include} the neuron detecting the body of a panda, which is also a very important feature to classify a panda.

\noindent \textbf{\normalsize R2 - Providing an overview of the datapath.}
In a large CNN, a datapath often contains millions of neurons and connections.
Directly presenting all neurons in a datapath will induce severe visual clutter.
Thus, it is necessary to provide experts an overview of a datapath.
E$_1$ commented, ``I cannot examine all the neurons in a datapath because there are too many of them.
In the examining process, I often start by selecting an important layer based on my knowledge, and examine the neurons in that layer to analyze the learned features and the activations of these neurons.
%working mechanism of an adversarial example.
The problem of this method is when dealing with a new architecture, I may not know which layer to start with.
Thus, I have to examine a bunch of layers, which is very tedious.''
Thus, it is necessary to provide the experts an overview of the datapath with visual guidance to facilitates experts in selecting the layer of interest.
% Mengchen comment: structure not enough?
% A traditional way is to visualize the structure of 
% This will facilitate experts to find 
The requirement of providing an overview of a CNN aligns well with previous research~\cite{kahng2018acti,liu2018analyzing,wongsuphasawat2018visualizing}.

\noindent \textbf{\normalsize R3 - Exploring a datapath at the detailed levels.}
Although the overview of a datapath facilitates experts in finding the layer of interest, it is not enough to diagnose the root cause of the wrong prediction introduced by an adversarial example.
The experts said that they wanted to link the overview of a datapath with detailed neuron activations.
This linkage helps them identify the most important neurons that lead to the misclassification.
%Thus, it is necessary to allow experts to explore a datapath from high-level layers to detailed neurons activations.
Since a layer may contain millions of critical neurons, the experts also desired a medium level between the layer level and neuron level.
For CNNs, the experts recommended to group neurons into feature maps.
%Thus, we added a feature map level as the medium level.
E$_2$ said that, ``Neurons in a feature map learn to detect the same feature.
%Their difference is that they learn to detect the same feature at different location of the image.
Grouping them is very common in our research.''
%Thus, we decided to allow experts explore datapath at three levels: layer-level, feature map level, and neuron level.
Previous research also indicates that visual analytics for deep learning benefits from multi-level visualization~\cite{kahng2018acti,liu2018analyzing}.
%\looseness=-1 

% at multiple levels, i.e., layer level, feature map level, and neuron level.
% However, a datapath may contains millions of neurons.
% Directly visualizing all neurons is prohibitive.
% The experts need a exploration mechanism enabling them to explore the current level, find the visual element of interest, and zoom into the next more detailed level.

% as a starting point of
% exploration, and let users switch to the detailed inspection of activations.

\noindent \textbf{\normalsize R4 - Comparing multiple datapaths.}
An adversarial example is often generated by slightly perturbing the pixel values of a normal image.
Accordingly, a normal image and the corresponding adversarial example are nearly the same in the input space.
However, their prediction results are different.
The experts are interested in how they \rev{diverge} to different predictions.
%For example, E$_2$ commented, ``I want to know whether there are some critical `diverging points' in a CNN or it happens gradually through the network.''
For example, E$_2$ commented, ``I want to know whether there are some critical `diverging points' for the prediction difference or it is accumulated gradually layer by layer through the network.''
To this end, E$_2$ desired to compare the datapaths of normal examples and adversarial examples.
\rev{Inspired} by E$_2$, E$_1$ added that it was interesting to compare the datapath of an adversarial example (e.g., a panda image that is misclassified as a monkey) with that of the images from the predicted class (e.g., normal images containing monkeys). 
Such comparison helps them understand how these very different images ``merge'' into the same prediction (e.g., the monkey). 
The need of visual comparison is consistent with the findings of previous research~\cite{alexander2016task, gleicher2018considerations,liu2014survey}.\looseness=-1

% Thus, comparing the datapaths of the adversarial example and the original normal image is useful for detecting critical layers and feature maps

\subsection{System Overview}

Driven by the requirements collected from the experts, we have developed a visual analytics tool, AEVis, to illustrate the root cause of the robustness issues caused by adversarial examples.   
%motivated us to develop AEVis, which consists of the following modules:
This tool consists of the following two parts. 
\begin{compactitem}
\item\noindent{\textbf{\normalsize A datapath extraction module} that extracts the critical neurons and their connections for a set of selected examples (\textbf{R1}).}
\item\noindent{\textbf{\normalsize A datapath visualization module} that provides an overall understanding of the datapath of interest (\textbf{R2}), illustrates datapaths in multiple levels (\textbf{R3}), and facilitates experts to visually compare several datapaths (\textbf{R4}).} 

%allow experts explore and compare datapaths from high-level layers to detailed neuron activations (\textbf{R2}, \textbf{R3}, \textbf{R4}).}
% Mengchen comment: whether to write this in more detail

% 
% \item\noindent{\textbf{\normalsize An interaction module} that provides a set of interactions such as comparing multiple datapaths (\textbf{R4}).}
\end{compactitem}

As shown in Fig.~\ref{fig:system}, AEVis takes a trained CNN and the examples to be analyzed as the input.
The examples include both normal examples (Fig.~\ref{fig:system}A) and adversarial examples (Fig.~\ref{fig:system}B).
% Given several examples sets of interest, 
% The examples includes both normal examples (Fig.~\ref{fig:system}A) and adversarial examples (Fig.~\ref{fig:system}B)
Given the examples and the CNN, the datapath extraction module extracts the critical feature maps and their connections that are responsible for the predictions of the examples (Fig.~\ref{fig:system} (a)).
The extracted datapaths are then fed into the datapath visualization module (Fig.~\ref{fig:system} (b)), which supports the navigation and comparison of the datapaths from the high-level layers to the detailed neuron activations.
% Mengchen comment: whether to add a paragraph to illustrate the basic workflow

\mengchen{A typical analysis workflow of AEVis is shown in Fig.~\ref{fig:teaser}.
An expert explores the image list (Fig.~\ref{fig:teaser} (a)) and selects several groups of images for further understanding and analysis.
After clicking on the `Analyze' button, AEVis first provides an overview of the datapaths at the layer level (Fig.~\ref{fig:teaser} (b)).
At the layer level, each rectangle represents a layer group and a dot plot is combined with each layer group to illustrate the similarities between/among the extracted datapaths of the layers in each layer group.
By examining the dot plots, the expert is able to detect layers of interest and further examines the feature maps in the layer ($L_C$ and $L_D$ in Fig.~\ref{fig:teaser}).
In each layer, the Euler-diagram-based design helps the expert focus on the share/unique feature maps of several datapaths.
% shows the overlapping relationship of the feature maps between/among the datapaths.
Aided by this design and the color coding (e.g., activations) of feature maps, the expert can select a feature map of interest ($F_C$ and $F_D$ in Fig.~\ref{fig:teaser}), and explore the detailed neuron activations as well as the learned features of the neurons in this feature map (Fig.~\ref{fig:teaser} (c)).
It facilitates him/her in finding the root cause (e.g., a set of neurons) for the misclassification of the adversarial examples.
}

% Experts can also interact with the generated visualization for further analysis.
% For example, experts can interactively select several groups of images and visually compare the datapaths of them.

\section{Datapath Extraction}
\label{sec:datapath}

\begin{figure}[!t]
  \centering
   \vspace{-3mm}
  \includegraphics[width=\linewidth]{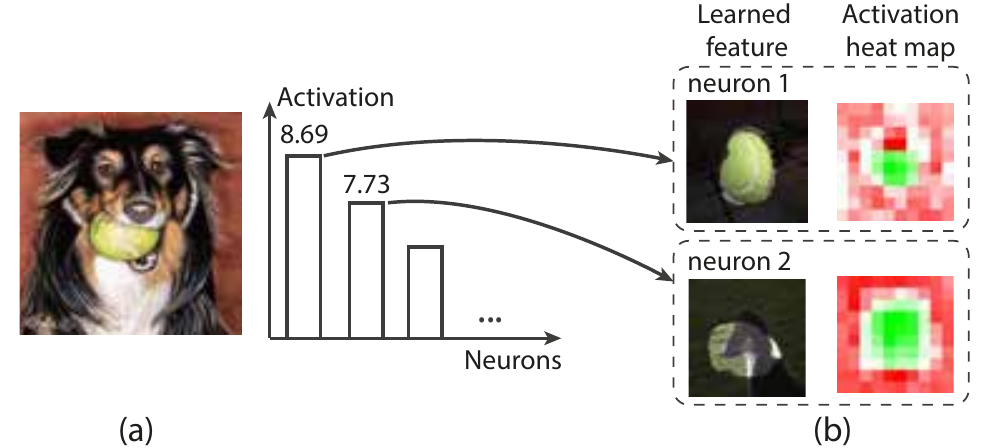}
  \caption{A misleading result of the activation-based datapath extraction approach: (a) the input image; (b) the top 2 critical neurons found by the activation-based approach.
  The learned feature is computed by the discrepancy maps~\cite{zhou2014object} and the activation heat map show the activations on the corresponding feature map.}
  %(c) crithttps://www.overleaf.com/11629216fdkdynbhtcqt#ical neurons found by our back-propagation based approach.}\looseness=-1
  \label{fig:datapath_extraction_moti}
  \vspace{-4mm}
\end{figure}

% \begin{figure}[!b]
%   \centering
% %   \vspace{-4mm}
%   \includegraphics[width=\linewidth]{fig/datapath_extraction_alg.pdf}
%   \caption{Datapath extraction algorithm pipeline}\looseness=-1
%   \label{fig:datapath_extraction_alg}
% \end{figure}

% Based on the characteristics of a CNN, we formulate the datapath extraction problem as a subgraph extraction problem and develop a  to solve the problem.

\subsection{Motivation and Problem Formulation}

% Datapath extraction aims to extract the critical neurons and their connections that are responsible for the prediction of a set of examples.
% It is the basis for analyzing why an adversarial example is misclassified.
Extracting the datapath that are responsible for the prediction of a group of examples
is the basis for analyzing why an adversarial example is misclassified (\textbf{R1}).
The key challenge is to identify the critical neurons as selecting the corresponding connections to form the datapath is straightforward.
% Next, we only discuss how to select critical neurons.
Next we discuss how to extract the critical neurons for an individual example and then extend our approach to selecting critical neurons for a set of examples.

%To select critical neurons for an example, the experts often treat the most activated neurons as the critical neurons~\cite{bilal2018convolutional}.
A commonly used method is to treat the most activated neurons as the critical neurons~\cite{bilal2018convolutional,kahng2018acti}.
However, this method may result in misleading results, especially when a highly recognizable secondary object is mixed with the main object in an image.
% or there are surrounding objects around the main object
For example, Fig.~\ref{fig:datapath_extraction_moti} shows the top 2 critical neurons found by such activation-based approach for a sheepdog image (Fig.~\ref{fig:datapath_extraction_moti} (a), employed network: ResNet-101~\cite{He2016_CVPR_residual}, image label: Shetland sheepdog).
We can see that the second highly activated neuron learns to detect the head of a dog (Fig.~\ref{fig:datapath_extraction_moti} (b)), which is indeed critical for classifying a sheepdog.
However, the most critical neuron learns to detect a ball (Fig.~\ref{fig:datapath_extraction_moti} (b)), which is not an important feature for classifying a sheepdog.
Such a misleading result root\rev{s} in that the neurons have complex interactions with each other and the activations of the neurons are processed by a highly nonlinear function to generate the final prediction.
Thus, highly activated neurons may not be the critical neurons for a prediction.
Such critical neurons are the neurons that highly contributed to the final prediction.
In other word, by only combining the contributions of the critical neurons, the prediction of an example will not be changed.
To this end, we aim at selecting a minimized subset of neurons, which keep the original prediction.
Accordingly, we formulate critical neurons extraction as a subset selection problem:
% \begin{equation}
% \label{eq:dp-ext1}
% N^{opt}=\underset{{N_s}\subseteq N}{\mathop{\arg \min }}\,(p(x)-p(x;{N_s}))^2,
% \end{equation}
% \begin{small}
\begin{equation}
\label{eq:dp-ext2}
N^{opt}=\underset{{N_s}\subseteq N}{\mathop{\arg \min }}\,(p(x)-p(x;{N_s}))^2 +\lambda|N_s|^2 .
\end{equation}
% \end{small}
The first term is to keep the original prediction and the second term ensures to select a minimized subset of neurons.
Specifically, $N$ is the set of the neurons in a CNN, $N_s$ is a subset of neurons $N$, $N^{opt}$ is the calculated critical neurons, $p(x)$ is the prediction of example $x$, and $p(x;N_s)$ is the prediction if we only consider the neuron subset $N_s$.
To measure the difference between two predictions, we adopt the widely used L2-norm.
$|N_s|$ is the size of $N_s$ and $\lambda$ is used to balance the two terms.
\rev{A larger $\lambda$ indicates a tendency to select a smaller subset of neurons.}
%\looseness=-1

% However, directly solving Eq.~\ref{eq:dp-ext1} will result in a trivial solution $N^{opt} = N$.
% Thus, we add a regularization term to favor small set of critical neurons instead of the whole set of neurons:

% \subsection{Datapath Aggregation}

As discussed in Sec.~\ref{sec:require}, extracting the common datapath for a set of examples instead of an individual example can improve the representativeness of the extracted datapath (\textbf{R1}).
A natural extension of Eq.~(\ref{eq:dp-ext2}) is minimizing the sum of the difference in the prediction of the example set $X$:   
\begin{equation}
\label{eq:dp-set}
N^{opt}=\underset{{N_s}\subseteq N}{\mathop{\arg \min }}\, \sum\limits_{x_k \in X}{(p(x_k)-p(x_k;{N_s}))^2 +\lambda|N_s|^2}.
\end{equation}
Next, we discuss how to effectively solve this problem.
For simplicity, we take Eq.~(\ref{eq:dp-ext2}) (critical neurons for one example) as an example to illustrate the solution.

\subsection{Solution to Subset Selection}

Directly solving the subset selection problem (Eq.~(\ref{eq:dp-ext2})) is time-consuming because: (1) it is an NP-complete problem and (2) the search space is prohibitively large due to the large number of neurons in a CNN.
% (2) the subset selection problem itself is .
We therefore combine a divide-and-conquer approach with a quadratic optimization to reduce the search space and find a more accurate approximation.
%To tackle these challenges, our optimization approach contains two steps.
% \begin{compactitem}
% \item\noindent{\textbf{\normalsize Search space reduction} that reduces the large search space by splitting the subset selection problem into a series of separate subproblems and grouping the neurons into feature maps.}
% \item\noindent{\textbf{\normalsize Quadratic approximation} that approximates each NP-complete subproblem as an easy-to-solve quadratic optimization by Taylor decomposition, and solve it based on back-propagation.} 
% % \textbf{Step 2}: because each generated subproblem is still NP-complete, we .
% \end{compactitem}
In particular, we develop a divide-and-conquer-based \textbf{search space reduction} method by splitting the subset selection problem into a series of separate subproblems and grouping the neurons into feature maps.
As each subproblem is still NP-complete, we then employ the \textbf{quadratic approximation} to more accurately approximate each NP-complete subproblem as an easy-to-solve quadratic optimization by Taylor decomposition\rev{~\cite{rudin1964principles}}, and solve it based on back-propagation\rev{~\cite{christopher2006_PRML}}.

% split the optimization problem into a set of sub-optimization problem and for each sub-optimization we derive a fast numerical optimization method based on back-propagation.

\subsubsection{Search space reduction}

A CNN is traditionally represented as a directed acyclic graph (DAG), where each node represents a neuron and each edge denotes a connection between nodes.
A widely-used approach to accelerate DAG-related algorithms is processing the nodes layer by layer~\cite{liu2013storyflow,sugiyama1981methods}.
\rev{Inspired} by these algorithms, we split the original subset selection problem (Eq.~(\ref{eq:dp-set})) into a set of subproblems.
Each selects the critical neurons in one layer:
$N^i_{opt}={\mathop{\arg \min }}\,(p(x)-p(x;{N_s^i \cup N^{-i}}))^2 +\lambda^i|N_s^i|^2,$
where ${N_s^i}\subseteq N^i$, $N^i$ is the set of the neurons in layer $i$, and $N^{-i}$ is the set of all other neurons in the CNN except the ones in layer $i$.
After solving all the subproblems, we aggregate all the sub-solutions $N^i_{opt}$ into the final critical neuron set $N_{opt} = \bigcup\limits_{i}{N^i_{opt}}$.
% \noindent \textbf{\normalsize Problem splitting.}
% Neurons $N$ in a CNN can be naturally divided into separate set of neurons  by their layer,  

% A CNN is composed of many different functions, each of which we call a layer.

% A widely used approach to accelerate is process layer by layer.
% Based on the same idea, we split the original XXX
% In particular, for a CNN with $L$ layers, we split , each of which selects the neurons in one layer:

% \begin{equation}
% \label{eq:dp-ext-layer}
% N^i_{opt}=\underset{{N_s^i}\subseteq N^i}{\mathop{\arg \min }}\,(p(x)-p(x;{N_s^i \cup N^{-i}}))^2 +\lambda^i|N_s^i|^2,
% \end{equation}

Although dividing the original problem into a set of subproblems can largely reduce the search space, the search space of each subproblem is still large because a layer may contain more than one million neurons.
Thus, we group neurons into a set of feature maps to further reduce the search space.
%In particular, in a CNN, neurons in a layer can be organized into feature maps.
In a CNN, neurons in a feature map share the same weights, and thus learn to detect the same feature.
%Utilizing this characteristics of a CNN, we group the neurons into feature maps:
Utilizing this characteristics, we formulate the feature map selection problem as
% to further reduce the search space:
\begin{equation}
\label{eq:dp-ext-fm}
F^i_{opt}=\underset{{F_s^i}\subseteq F^i}{\mathop{\arg \min }}\,(p(x)-p(x;{F_s^i \cup F^{-i}}))^2 +\lambda^i|F_s^i|^2,
\end{equation}
where $F$ is the set of feature maps in a CNN, $F^i$ is the set of the feature maps in layer $i$, $F_s^i$ is a subset of $F^i$, and $F^{-i}$ is the set of all other feature maps in the CNN except the ones in layer $i$. 

\subsubsection{Quadratic Approximation}
% \noindent \textbf{\normalsize Sub-optimization Solving.}
% \mengchen{NOT THIS ARGUMENT}
% To solve each sub-optimization (Eq.~\ref{eq:dp-split}), Ribeiro et al. propose a model-agnostic sampling method.
% Directly adopting their method is time-consuming for deep model.
% For example, their method will take around 10 minutes for solving this problem when applying in the Inception network \mengchen{ref}.
% To solve this problem, we utilize some model-dependent information (e.g., gradients and activations in a deep model) and develop a fast optimization method based on back-propagation.

Although we have reduced the search space from millions of dimensions (neurons in a layer) to thousands of dimensions (feature maps in a layer), it is still time-consuming to solve Eq.~(\ref{eq:dp-ext-fm}) because it is an NP-complete discrete optimization problem.
% which makes it a NP-complete problem.
To tackle this issue, we transform the discrete optimization into a continuous optimization problem.
% To further approximate the continuous optimization problem as an easy-to-solve quadratic optimization.
In particular, we reformulate Eq.~(\ref{eq:dp-ext-fm}) as:
% \begin{equation}
% \label{eq:dp-refor}
$\mathbf{z}_{opt}^i={\mathop{\arg \min }}\,(p(x)-p(x;\mathbf{z}^i))^2 + \lambda^i(\sum_{j}{z^i_j})^2,$
% \end{equation}
%\underset{z^i_j\in \{0,1\}}
% $\mathbf{z}_{opt}^i={\mathop{\arg \min }}\,(p(x)-p(x;\mathbf{z}^i))^2 + \lambda^i(\sum\limits_{j}{z^i_j})^2,$
where $\mathbf{z}^i = [z_1^i, ... , z_n^i]$ and $z_j^i \in \{0,1\}$ is an indicator to represent whether the $j$-th feature map in layer $i$ is critical.
If the feature map is critical, $z_j^i=1$, otherwise, $z_j^i=0$.
\rev{Inspired} by spectral clustering~\cite{ng2002spectral}, we approximate the discrete optimization with a continuous optimization by removing the discreteness condition $z_j^i \in \{0,1\}$ and allowing $z^i_j$ to take a value in $[0,1]$:
% substituting $z^i_j\in \{0,1\}$ with $z^i_j \in [0,1]$:
\begin{equation}
\label{eq:dp-cont}
\mathbf{z}_{opt}^i=\underset{z^i_j\in [0,1]}{\mathop{\arg \min }}\,(p(x)-p(x;\mathbf{z}^i))^2 + \lambda^i(\sum\limits_{j}{z^i_j})^2,
\end{equation}
Eq.~(\ref{eq:dp-cont}) can be solved using gradient-based numerical optimization approaches such as the BFGS (Broyden$-$Fletcher$-$Goldfarb$-$Shanno) algorithm~\cite{wright1999numerical}.
% Eq.~\ref{eq:dp-cont}.
The gradient ${\partial p}/{\partial z_j^i}$ is calculated by back-propagation~\cite{christopher2006_PRML}.
However, this method is computationally expensive because the gradient-based optimization is an iterative process where we have to calculate the gradients ${\partial p}/{\partial z_j^i}$ at each iteration.
%Computing the gradients for very deep CNNs (e.g., ResNet~\cite{He2016_CVPR_residual}, 50-200 layers) itself is time-consuming because 
According to the calculation process of back-propagation, the deeper a CNN is, the longer it takes to compute the gradients.

% Such an approach works well for medium deep CNNs, such as the VGG net~\cite{simonyan2014very} (15-20 layers).
% However, we find that for the state-of-the-art CNNs, such as ResNet~\cite{He2016_CVPR_residual} (50-200 layers), it takes more than \mengchen{5} seconds to compute the critical neurons in each layer, which sums up to \mengchen{more than 500 seconds to calculate the critical neurons for one example with 101 layers}.
% The reason is that the gradient-based optimization is an iterative process, where at each iteration we have to calculate the gradients $\frac{\partial p}{\partial z_j^i}$.
% According to back-propagation, the deeper a CNN is, the longer it takes to compute the gradients.
% Thus, for very deep CNNs, directly solving Eq.~\ref{eq:dp-cont} is computationally expensive.

To tackle this issue, we approximate Eq.~(\ref{eq:dp-cont}) as a quadratic optimization where we calculate the gradients only once.
Since $p(x) = p(\mathbf{a}_1,...,\mathbf{a}_n)$ and $p(x;\mathbf{z}^i) = p(z_j^i\mathbf{a}_1,...,z_j^n\mathbf{a}_n)$,
we rewrite Eq.~(\ref{eq:dp-cont}) as a multivariate function.
Here $\mathbf{a}_j$ is the activation vector of the $j$-th feature map produced by example $x$.
%Such reformulation indicates that $p(x)$ can be \shixia{regarded} as a multivariate function.
 %It triggers us to use Taylor decomposition~\cite{rudin1964principles} to decompose $p(x) - p(x;\mathbf{z}^i)$ into a linear form: 
We then the Taylor decomposition~\cite{rudin1964principles} is to decompose $p(x) - p(x;\mathbf{z}^i)$ into a linear form: 
$p(x) -p(x;\mathbf{z}^i) \approx \sum\limits_{j}{ (1-z_j^i) \mathbf{a}_j \cdot  {{\left. \frac{\partial p}{\partial \mathbf{a}_j^i} \right|}_{\mathbf{a}}} } $, where $\mathbf{a} = [\mathbf{a}_1^i, ..., \mathbf{a}_n^i]$ and $\mathbf{x}\cdot \mathbf{y}$ represents dot product between vectors $\mathbf{x}$ and $\mathbf{y}$.
% Similarly, $p(x;\mathbf{z}^i)$ can be decomposed into: $p(x;\mathbf{z}^i) \approx \sum\limits_{j}{ z_j^i \frac{\partial p}{\partial \mathbf{a}_j^i} \cdot \mathbf{a}_j } $.
By substituting the above decomposition into Eq.~(\ref{eq:dp-cont}), we obtain a quadratic optimization:
% \begin{eqnarray}
% \label{eq:dp-final}
% \mathbf{z}_{opt}^i & = & \underset{z^i_j\in [0,1]}{\mathop{\arg \min }}\,(\sum\limits_{j}{ (1-z_j^i) \mathbf{a}_j \cdot  {{\left. \frac{\partial p}{\partial \mathbf{a}_j^i} \right|}_{\mathbf{a}}} })^2 + \lambda^i(\sum\limits_{j}{z^i_j})^2\\
% \label{eq:dp-final2}
% & = & \underset{z^i_j\in [0,1]}{\mathop{\arg \min }}\, \mathbf{z}^i(Q^i+ \lambda^i I)(\mathbf{z}^i)^T -2q\mathbf{q}^i \cdot \mathbf{z}^i,
% \end{eqnarray}
\begin{equation}
\label{eq:dp-final2}
\mathbf{z}_{opt}^i = \underset{z^i_j\in [0,1]}{\mathop{\arg \min }}\, \mathbf{z}^i(Q^i+ \lambda^i I)(\mathbf{z}^i)^T -2q\mathbf{q}^i \cdot \mathbf{z}^i,
\end{equation}
%here I delete q^2 for simplyfying the next eq for set solution
where $\mathbf{q}^i = [\mathbf{a}_1^i \cdot {{\left. \frac{\partial p}{\partial \mathbf{a}_1^i} \right|}_{\mathbf{a}}}, \cdots , \mathbf{a}_n^i \cdot {{\left. \frac{\partial p}{\partial \mathbf{a}_n^i} \right|}_{\mathbf{a}}}]$, $q$ is the sum of $\mathbf{q}^i$, $Q=(\mathbf{q}^i)^T\mathbf{q}^i$ is a $n \times n$ matrix, and $I$ is a $n \times n$ identity matrix.
Solving Eq.~(\ref{eq:dp-final2}) only needs to evaluate the gradients once.
%and can be calculated by back-propagation.
We use the BFGS algorithm to solve Eq.~(\ref{eq:dp-final2}).
To control the scale of the two terms, the parameter $\lambda^i$ is set to $0.1/|F^i|^2$, where $|F^i|$ is the number of the feature maps in layer $i$.\looseness=-1

In the same way, we obtain a solution for selecting critical feature maps for a set of examples (Eq.~(\ref{eq:dp-set})).
Compared with Eq.~(\ref{eq:dp-final2}), there are two differences: (1) replacing $Q_i$ by $\sum{Q_k^i}$ and (2) replacing $q\mathbf{q}^i$ by $\sum{q_k\mathbf{q}_k^i}$, where $k$ is the index of the example in the example set.

% \begin{equation}
% \label{eq:dp-cont}
% G^{opt}_i=\underset{I_i^j\in [0,1]}{\mathop{\min }}\,L(P(x;G),P(x;I_i)) + ||I_i||_1,
% \end{equation},
% Here $z_i^j$ represents the importance of $j$-th component in layer $i$ in the datapath.
% It means we \mengchen{element-wisely} multiply $I_i^j$ with the corresponding activations.
% \mengchen{NEED TO MENTION SETTTING ACTIVATION TO ZERO BEFORE}
% If $L()$ is differentiable, we can use gradient-based numerical optimization approaches (e.g., BFGS) to solve
% the converted optimization.
% % Eq.~\ref{eq:dp-cont}.
% The gradient can be efficiently calculated by back-propagation.

% \noindent \textbf{\normalsize Example and Results.}
% Here we also use CNN as an example to demonstrate the optimization process and results.
% As mentioned before, we treat the feature maps in a layer as the components of the layer.
% Given an example $x$, the importance $I$ is implemented by 
% The function $L(a,b)$ is chosen to be the widely used L2-norm: $L(a,b) = (a-b)^2$.

%C. Szegedy, W. Liu, Y. Jia, P. Sermanet, S. Reed, D. Anguelov, D. Erhan, V. Vanhoucke, and A. Rabinovich. Going deeper with convolutions. In Computer Vision and Pattern Recognition (CVPR), 2015.

\section{Datapath Visualization}

% \subsection{Overview}
An extracted datapath usually contains millions of neurons and even more connections, 
%which prohibits experts to efficiently examine neurons one by one.
which prohibits experts \doc{from} efficiently examining the datapath layer by layer.
%to
To help experts effectively explore the extracted datapaths, we design a multi-level datapath visualization, which enables experts to analyze and compare datapaths from high-level layers to detailed neuron activations (\textbf{R2}, \textbf{R3}, \textbf{R4}).
Based on \doc{our} discussions with the experts, we visualize datapaths \doc{on} three levels: the layer level, \doc{the} feature map level, and the neuron level.
%from
At each level, we (1) calculate the layout of the items (layers, feature maps, neurons) to reveal the relationships among them; and (2) provide visual hints to guide experts in finding the item(s) of interest and zooming in to a more-detailed level (e.g., from the layer level to the feature map level).
%Next, we introduce the layout algorithm and the designed visual hints for each level.

% \begin{figure}[!t]
%   \centering
% %   \vspace{-4mm}
%   \includegraphics[width=\linewidth]{fig/dp_vis_overview.pdf}
%   \caption{\mengchen{TODO}}
%   %(c) critical neurons found by our back-propagation based approach.}\looseness=-1
%   \label{fig:dp_vis_overview}
% \end{figure}

\begin{figure}[htbp]
  \centering
%    \vspace{-2mm}
  \includegraphics[width=\linewidth]{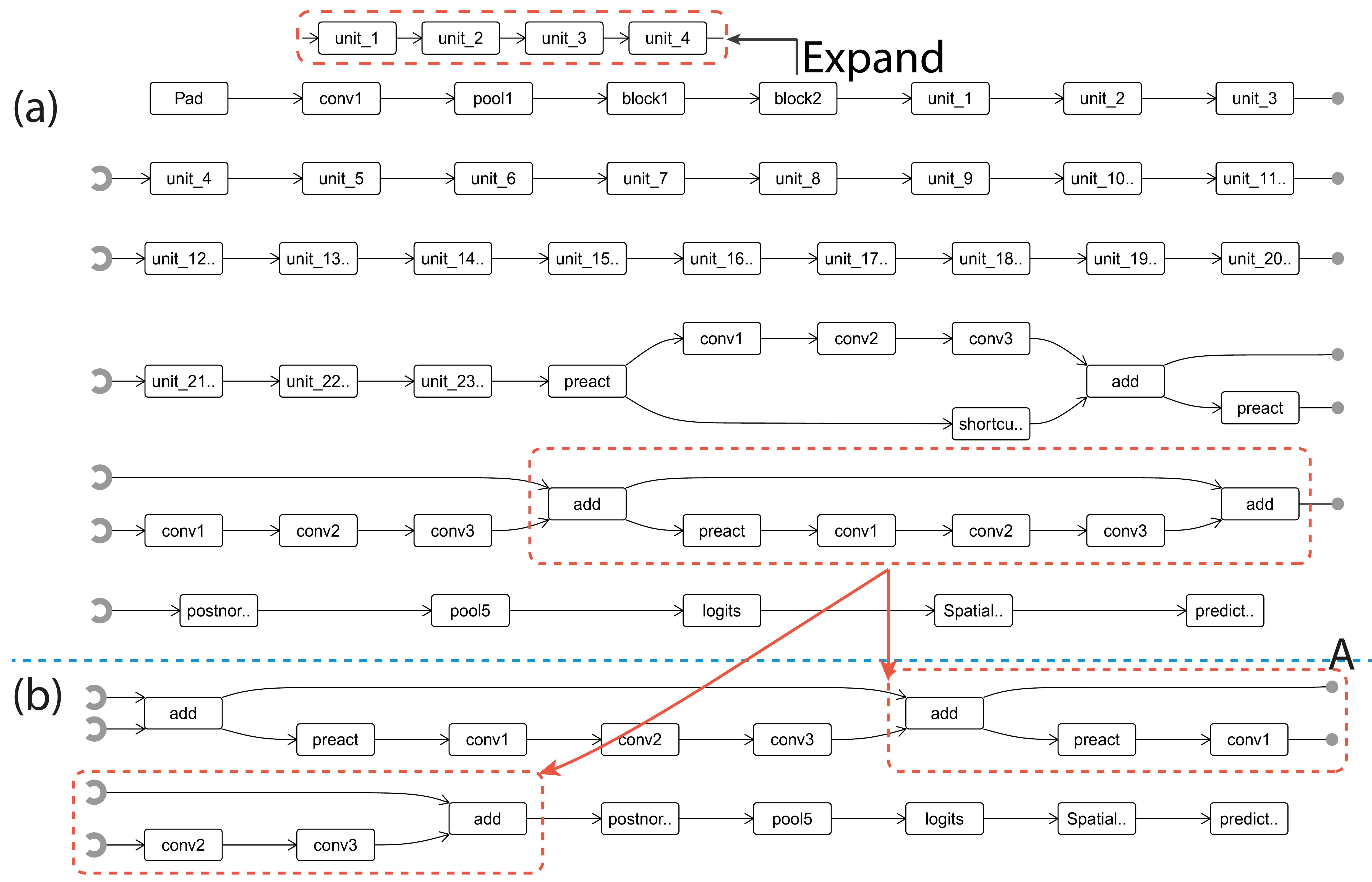}
  \caption{Segmented DAG visualization: (a) showing 49 layers of ResNet101 simultaneously; (b) the segmentation result calculated by only considering the empty space at the end of each segment.}
  \vspace{-4mm}
  %(c) critical neurons found by our back-propagation based approach.}\looseness=-1
  \label{fig:layer_vis}
\end{figure}

\subsection{Layer-level Visualization}

The layer-level visualization provides an overview of the extracted datapath, and guides experts in selecting a layer to examine (\textbf{R2}).

\noindent \textbf{\normalsize Layout.}
At the layer level, we focus on illustrating how layers are connected.
\rev{Initially, we employ the widely-used TensorFlow Graph Visualizer~\cite{wongsuphasawat2018visualizing}.}
In particular, we formulate a CNN as a DAG (directed acyclic graph), where each layer is a node and the connections between layers are edges.
Based on this formulation, a node-link diagram \rev{with a vertical layout} is employed to visualize the layers and their connections.
Then the DAG layout algorithm in\rev{~\cite{sugiyama1981methods}} is employed to calculate the position of each layer.
To handle large CNNs, a hierarchy of the layers is built, where each leaf node is a layer and each non-leaf node is a layer group~\cite{wongsuphasawat2018visualizing}.

\rev{The experts were overall satisfied with this design, but after they tried the prototype they commented that they often need to zoom in to the lower levels of the layer hierarchy in order to analyze the root cause of a misclassification.
It is tedious for them to zoom in for each adversarial example.
In addition, when they zoom in to the lower levels, the visualization often suffers from a long skinny strip with a very high aspect ratio (e.g., Fig. 8 (a) in~\cite{wongsuphasawat2018visualizing}).
This phenomenon is worse when they analyze state-of-the-art CNNs, such as ResNet~\cite{He2016_CVPR_residual} with 50-200 layers.}\looseness=-1 
%when they zoom in to the lower levels of the layer hierarchy 

\rev{To solve the above problems, we combine a treecut algorithm and a segmented DAG visualization to save experts' efforts and generate a layout with a better aspect ratio, respectively.}

\noindent{\textit{Treecut.}} \rev{To save experts' efforts of zooming in,} we use a treecut algorithm~\cite{cui2014hierarchical} to select an initial set of layers (around 50 layers) \rev{from the layer hierarchy}. 
In this algorithm, the DOI measures the datapath difference between two sets of images (e.g., adversarial examples and normal examples).

\noindent{\textit{Segmentation.} Inspired by the segmented timeline~\cite{brehmer2017timelines}, we propose \rev{transposing the vertical layout into a horizontal layout}, segmenting the initial DAG into several parts, and visualizing the segmented parts from top to bottom.
\rev{The experts commented that this horizontal design resembles a calendar and thus is familiar to them.}
%To maintain this familiarity, we use a horizontal layout instead of the vertical layout in~\cite{wongsuphasawat2018visualizing}.}
As shown in Fig.~\ref{fig:layer_vis} (a), our segmented DAG visualization effectively illustrates the connections among dozens of layers with good aspect ratios.

% \textit{Segmentation.}
% \rev{Inspired} by ,
The key challenge of segmenting a DAG is to decide where to segment it.
We formulate the segmentation problem as a ``printing neatly'' problem~\cite{cormen2009introduction}, \doc{in which} we aim to minimize the cost function that sums up the empty space at the end of each line while ensuring that no word is \doc{off} screen.
% out of the 
This optimization method is suitable for a CNN with a chain structure, such as VGG net~\cite{simonyan2014very}.
However, state-of-the-art CNNs (e.g., ResNet~\cite{He2016_CVPR_residual}, and DenseNet~\cite{huang2017densely}) often contain basic building blocks, whose layers can split and merge (e.g., Fig.~\ref{fig:layer_vis}A).
These building blocks are often connected with others as a chain.
Directly minimizing the above cost function may cut the basic building blocks apart, which hinders the understanding of the network structure \rev{(Fig.~\ref{fig:layer_vis} (b))}.
To solve this problem, we add a regularization term to the cost function to penalize a segmentation scheme that cuts a building block into two parts:
\begin{equation}
%\vspace{-2mm}
\label{eq:segment}
\underset{e_i \geq 0}{\mathop{\min }}\, \sum\limits_{i=1}^{k-1}{e_i + \lambda c_i},
%\vspace{-2mm}
\end{equation}
where $k$ is the number of segments, $e_i$ is the empty space at the end of segment $i$, $c_i$ represents whether a building block is cut by segment  $i$, and $\lambda$ is used to balance the two terms.
In AEVis, experts can interactively modify $\lambda$.
% This term is used to favor segmentation result that cuts outside basic building blocks.
Eq.~(\ref{eq:segment}) can be efficiently solved using dynamic programming.
As shown in Fig.~\ref{fig:layer_vis}, if we only consider the empty-space objective, \doc{building block A} will be cut into two parts (Fig.~\ref{fig:layer_vis} (b)).
%doc remove the
Adding the regularization term can avoid unnecessary cuts by balancing between the small empty space and the protection of the building blocks (Fig.~\ref{fig:layer_vis} (a)).

\begin{wrapfigure}{r}{0.15\textwidth}
\vspace{-0.15in}
\includegraphics[width=0.15\textwidth]{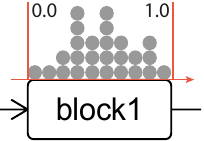}
%,trim=-0.4cm 0cm -0.4cm 0cm
\vspace{-0.2in}
\caption{A dot plot as the visual hint for layer selection. }
\label{fig:dot}
% \vspace{-0.1in}
\end{wrapfigure}
\noindent \textbf{\normalsize Visual hint.}
Showing how the layers are connected is not enough \doc{in} guiding experts to find the layer of interest.
%to
Thus, for each node (a layer or layer group) in the segmented DAG, we provide a visual statistical description of the datapath(s) in that layer (group).
% there is limited space for each node, we need a compact visualization.
Because there \doc{is} limited space for each node and a layer group usually contains dozens of layers, we employ a dot plot as the visual hint, which is a compact visualization and widely used for relatively small datasets\rev{~\cite{wilkinson1999dot}}.
In particular, in a dot plot, each dot represents \rev{a high-level statistics of a layer} (Fig.~\ref{fig:dot}).
The position of a dot on the x-axis denotes \rev{the value of the high-level statistics} of the datapath(s) in that layer, such as the activation similarity between two datapaths (the datapath for adversarial examples and the one for normal examples).
\rev{In particular, the activation similarity is calculated as $\sum{sim(I_A^i,I_N^i)}/N$, where $sim(.,.)$ is the widely-used cosine similarity, $I_A^i, I_N^i$ is a pair of adversarial and normal examples, and $N$ is the number of such pairs in the examples.
Other high-level statistics include the averaged activations of an example set on the corresponding datapath in that layer and the topological similarity between two datapaths (measured by Jaccard similarity).}
Examining the dot plots node by node helps experts detect the ``diverging point,'' where a normal image and its corresponding adversarial example diverge into different predictions (Fig.~\ref{fig:case0_overview}).
\rev{Experts are able to examine the feature maps in the layer group of interest (Fig.~\ref{fig:teaser}) or expand it to examine child layers or layer groups (Fig.~\ref{fig:layer_vis} (a)).}

% In AEVis, we provides several , such as 
% Thus, dot plot can both help experts have an overview of the size of the layer group and also get an overview. 
% With such overview, experts can determine the diverging point comparing datapaths.
% For example, \mengchen{TODO case}

\subsection{Feature-map-level Visualization}

%where
% \begin{figure}[t]
%   \centering
% %   \vspace{-4mm}
%   \includegraphics[width=0.85\linewidth]{fig/fm_vis.pdf}
%   \vspace{-2mm}
%   \caption{\todo{TODO:} Two alternative designs of feature-map-level visualization: (a) a pie chart is utilized to encode the belonging datapaths of a feature map; (b) a treemap is employed to encode the relationships among datapaths.}
%   %(c) critical neurons found by our back-propagation based approach.}\looseness=-1
%   \label{fig:fm_vis}
%   \vspace{-4mm}
% \end{figure}

When an expert detects a layer of interest, he/she then zooms in to the layer to examine the critical feature maps in that layer.
To preserve the analysis context, we visualize the feature maps in the selected layer as \doc{the} focus and other layers are shown as context. (Fig.~\ref{fig:teaser} (b)).

\noindent \textbf{\normalsize Layout.}
The belonging relationships between the feature maps and the extracted datapaths in one layer, including the unique feature maps of each datapath and the shared ones between/among them, are very useful for understanding the roles of these feature maps. 
\rev{As finding unique/share elements based on their set (datapath) membership is an important task tackled by the set visualization~\cite{alsallakh2014visualizing}, we then formulate the feature map layout problem as a set visualization problem.}
%In AEVis, each datapath is represented as a set and each feature map is an element that belongs to one or more sets (datapaths).
%In a Euler diagram, sets are represented by closed curves (e.g. rectangles) and the relationships among sets are illustrated by the curve overlaps.
\rev{Among various set visualization techniques, such as the Euler diagram, the line-based techniques (e.g., LineSets~\cite{alper2011design}), and the matrix-based techniques (e.g., ConSet~\cite{kim2007visualizing}), we decide to employ on the Compact Rectangular Euler Diagram~\cite{riche2010untangling} (ComED) because:
\begin{compactitem}
\item It can well depict set relations and thus disclose the unique/share feature maps~\cite{alsallakh2014visualizing};
\item Machine learning experts are familiar with Euler diagrams and they often use them to understand the set relationships;
%\item The scalability is suitable in our scenario because experts usually compare no more than four datapaths at the same time;
\item The number of feature maps in each datapath is clearly conveyed in the Euler diagram, which is important for the analysis. 
\end{compactitem}}
%\rev{In AEVis, the advantages of the Euler diagram over other set visualization techniques  are as follows:
% because approximately four objects can be tracked in a visual comparison~\cite{intriligator2001spatial,wang2016topicpanorama,yantis1992multielement};

% To improve the scalability and readability of Euler diagrams, researchers have proposed several variants of the Euler diagram~\cite{micallef2012assessing,riche2010untangling}.
% After analyzing these variants, we decide to employ the ~\cite{riche2010untangling} (ComED) because the layout result is compact and thus suitable for examining the feature maps in multiple layers simultaneously (Fig.~\ref{fig:teaser}L$_C$ and L$_D$).

%Ori: split sets with intersections to produce a strict hierarchy which can be easily drawn with non-overlapping convex shapes (groups) and then to link up the split regions with lines
\rev{As in ComED~\cite{riche2010untangling}, we split the datapaths in the layer of interest by their intersection relations.
This produces a hierarchy, which is visualized by non-overlapping rectangles (Fig.~\ref{fig:teaser}L$_C$).
Then the split parts of datapaths are connected with lines, which can be shown on demand.
Compared with ComED, we make two modifications to reduce visual clutter and better facilitate experts' visual comparison:
\begin{compactitem}
\item Utilize K-Means~\cite{christopher2006_PRML} to cluster the feature maps by their activations to reduce the visual clutter caused by a large number of feature maps (Fig.~\ref{fig:fm_vis_hint}).
Experts can interactively modify the parameters of the clustering (e.g., the number of the clusters $k$) to reduce the parameter sensitivity.
% Based on the clustering results, we arrange the feature map clusters on a grid in a treemap cell. 
% Experts can expand a feature map cluster of interest and examine the corresponding feature maps.
\item Use the Treemap layout instead of the graph layout because the graph layout result has an irregular boundary, which is not suitable for juxtaposing multiple layers to compare the datpaths in them (Fig.~\ref{fig:teaser}L$_C$ and L$_D$).
% \item Show the the connection lines on demand to reduce visual clutter caused by the lines connecting multiple parts,.
\end{compactitem}}

\begin{wrapfigure}{r}{0.15\textwidth}
\vspace{-0.15in}
\includegraphics[width=0.15\textwidth]{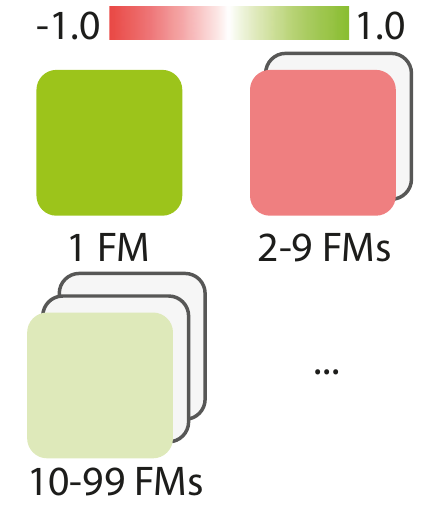}
%,trim=-0.4cm 0cm -0.4cm 0cm
\vspace{-0.2in}
\caption{The visual hints for a feature map cluster. FM is short for feature map.}
\label{fig:fm_vis_hint}
% \vspace{-0.1in}
\end{wrapfigure}
\noindent \textbf{\normalsize Visual hint.}
Although the \rev{Euler-diagram}-based design reveals the shared/unique feature maps very well, it  \doc{does not significantly help} to efficiently select an individual feature map cluster of interest for further examination.
%helps little
To this end, for each cluster, we illustrate its average importance ($z_j^i$ in Eq.~(\ref{eq:dp-cont}), Sec.~\ref{sec:datapath}) or average activation, because experts usually start their analysis from the most critical or most highly activated feature maps. 
The averaged importance/activation of a feature map cluster is encoded by the color of the feature map cluster (Fig.~\ref{fig:fm_vis_hint}).
In addition to importance/activation, we also encode the size of a feature map cluster because it is a basic information for a cluster.
%Ori: We employ a series of stacked rectangles to represent a feature map cluster (Fig.~\ref{fig:fm_vis_hint}).
% Triggered by Squares~\cite{ren2017squares}, we employ , which is proportional to the log of the cluster size. 
\rev{We employ a series of stacked rectangles to represent a feature map cluster (Fig.~\ref{fig:fm_vis_hint}), and use the number of stacked rectangles to encode the size of a cluster. 
To avoid the visual clutter caused by a large number of stacked rectangles, the number of rectangles $N_R$ is proportional to the log of the cluster size $N_C$: $N_R=log_{10}(N_C)+1$.}
For \doc{the sake of} consistency, a cluster with only one feature map in it is shown as a single rectangle.\looseness=-1

% In AEVis, we provide two methods, such as the average activation of the feature map cluster and the averaged importance score $z_j^i$  (Sec.~\ref{sec:datapath}) in the cluster.

\subsection{Neuron-level Visualization}
When an expert finds a feature map of interest, AEVis allows him/her to examine the neurons in that feature map.
As there are hundreds or even thousands of neurons in a feature map and dozens of images are often analyzed simultaneously, we cannot show all the activations in place due to the limited space.
Thus, we add a neuron panel (Fig.~\ref{fig:teaser} \rev{(c)}) to show the neurons.
To preserve the visual link~\cite{Steinberger2011TVCG_link} of the selected feature map and the corresponding neurons, we add the same label to the feature map and the neurons (Fig.~\ref{fig:neuron_vis}A).  

\noindent \textbf{\normalsize Layout.}
In a feature map, the neurons are naturally organized on a grid.
%COPY from DGM
\mengchen{The position of a neuron is determined by the position of the image patch that influences the activation of this neuron~\cite{goodfellow2016deep}.}
%Accordingly, we arrange the neurons of a feature map on a grid.

\noindent \textbf{\normalsize Visual hint.}
Following previous research~\cite{bilal2018convolutional,liu2017towards},  we employ the learned features of neurons and the activations of the neurons to help expert understand the roles of the neurons for the prediction.
%The necessity of these two facets of neurons are consistent with the findings of .

The activation of a neuron is represented by its color (red: negative activation; green: positive activation).
Combining the color coding and the grid-based layout of neurons, experts can detect which part of the image highly activates the neurons in the feature map.
For example, in Fig.~\ref{fig:neuron_vis}, we can find the neurons on the top right corner are highly activated, which corresponds to the panda head in the image.\looseness=-1
%For examples, in Fig.~\ref{fig:datapath_extraction_moti} (b), we can find the middle feature map 

To visualize the learned features of neurons, we first try the method used in previous \doc{papers} on explainable deep learning~\cite{bilal2018convolutional,liu2017towards}.
This method selects the image patches that highly activate a neuron to represent its learned feature.
%works
% and datasets with large image size (e.g., ImageNet, 224 pixels*224 pixels)
However, we discovered that when handling very deep CNNs (e.g., ResNet101~\cite{He2016_CVPR_residual}), this method cannot illustrate the \mengchen{exact} region in each image patch that highly activates a feature map neuron, especially for the neurons in top layers.
\doc{This is because} the activations of neurons in top layers of a very deep CNN are influenced by a large image patch.
 %The reason is that
For example, the neurons in more than \rev{at layer 10 and deeper} are influenced by all the pixels in an image in ResNet101.
%\doc{half the layers}
Thus, we employ the discrepancy map~\cite{zhou2014object} to highlight which region in the image actually highly activates a neuron and treat this region as the learned feature of this neuron.
To calculate the discrepancy map for a neuron, Zhou et al.~\cite{zhou2014object} occluded a very small patch of an image (8 pixels$\times$ 8 pixels), and calculated the new activation of the neuron produced by the occluded image.
If the activation changes \doc{a great deal}, the small patch is marked as important.
%a lot
This process iterates many times, and \doc{for} each iteration, a different patch is checked.
After this process, all the important small patches of the image \doc{remain} unchanged and other pixels are deemphasized by lowering their lightness.
%keep
For example, in Fig.~\ref{fig:neuron_vis}, by examining the discrepancy maps, we \doc{found} that the neurons in the selected feature map learn\doc{ed} to detect a panda head.
Without detecting the important region, we may draw \doc{the} wrong conclusion that the neurons learn\doc{ed} to detect a whole panda. 
%a
% An interesting discrepancy map is the third one, which contains a black-and-white dog.
% Highlighting the important region that actually highly activates a feature map 
%For example, in Fig.~\ref{fig:dp_extract_result_single} (b), we 

\begin{figure}[htbp]
  \centering
  \vspace{-1mm}
  \includegraphics[width=0.7\linewidth]{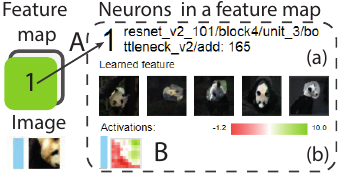}
  \vspace{-2mm}
  \caption{Neuron-level visualization: (a) the learned feature is shown as discrepancy maps; (b) activations are shown as a heat map.}
  \vspace{-2mm}
  %(c) critical neurons found by our back-propagation based approach.}\looseness=-1
  \label{fig:neuron_vis}
\end{figure}

\section{Evaluation}

\doc{As part of our evaluation}, we first performed a qualitative evaluation to demonstrate the effectiveness of the datapath extraction algorithm.
Then, two case studies were conducted to illustrate how AEVis helps \rev{the} experts \rev{E$_1$ and E$_2$} analyze both white-box and black-box adversarial examples. 

\subsection{Qualitative Analysis of Datapath Extraction}
\label{sec:eval_dp}

\begin{figure}[htbp]
  \centering
%   \vspace{-2mm}
  \includegraphics[width=\linewidth]{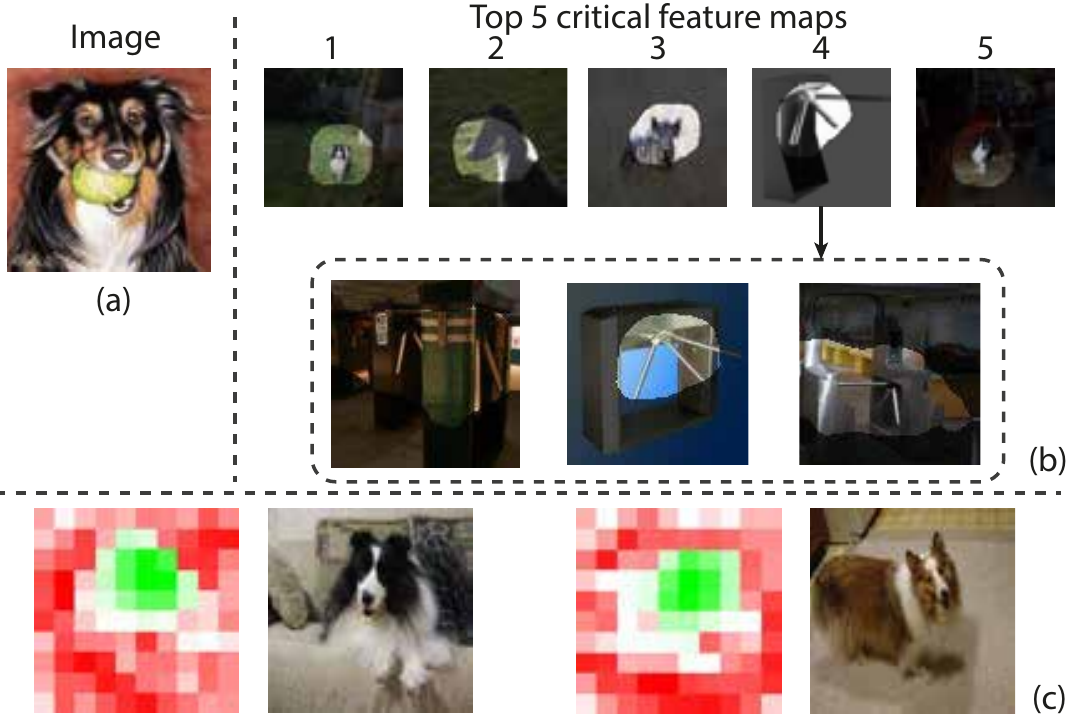}
  \vspace{-4mm}
  \caption{Critical feature maps extracted by our approach for a sheepdog image: (a) the input image; (b) top 5 the critical feature maps; (c) activation heat maps of \rev{the} neurons in feature map 4 on two examples.}
  %(c) critical neurons found by our back-propagation based approach.}\looseness=-1
  \label{fig:dp_extract_result_single}
\end{figure}

\begin{figure}[htbp]
  \centering
%   \vspace{-2mm}
  \includegraphics[width=\linewidth]{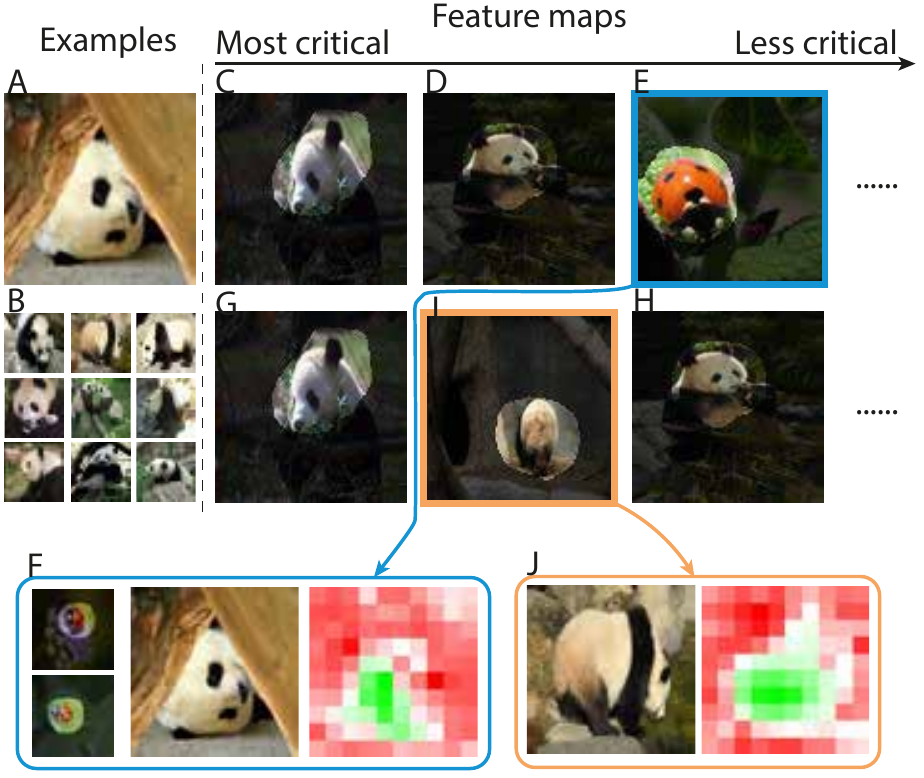}
  \vspace{-4mm}
  \caption{Comparison between the extracted datapath for one example and a set of examples.}
  \vspace{-5mm}
  %(c) critical neurons found by our back-propagation based approach.}\looseness=-1
  \label{fig:dp_extract_set}
\end{figure}

\noindent \textbf{\normalsize Datapath extraction for a single example.}
Fig.~\ref{fig:dp_extract_result_single} shows the top 5 critical feature maps extracted by our approach for the same image in Fig.~\ref{fig:datapath_extraction_moti}.
Feature maps 1,2,3 and 5 (Fig.~\ref{fig:dp_extract_result_single} (b)) learn to detect the features of dog ears, head, etc.
These features are indeed important to classify a Shetland sheepdog.
Compared with the activation-based approach, our approach is able to ignore the irrelevant feature map that learns to detect a ball (Fig.\ref{fig:datapath_extraction_moti} (b)).
Besides these easy-to-understand \rev{feature maps}, the learned feature of feature map 4 (Fig~\ref{fig:dp_extract_result_single} (b)) is difficult to understand at first glance.
To understand what this feature map \doc{actually} learns, we calculate a set of discrepancy maps (Fig~\ref{fig:dp_extract_result_single} (b)).
% for this feature map
A common property of these discrepancy maps is that they have white stripes with black strokes.
Thus, we conclude that this feature map learns to detect such \doc{a} feature.
However, it is still unclear why this feature map is considered critical in classifying a sheepdog.
To answer this question, we loaded more sheepdog images.
%We found that a distinctive feature of a sheepdog was that it has a white strip between its two dark-colored eyes.
We found that a distinctive feature of a sheepdog was a white strip between its two dark-colored eyes.
This feature was ignored by us at first because it is not obvious in the original example (Fig.~\ref{fig:dp_extract_result_single} (a)).
To verify our assumption, we further visualized the activations on the feature map (Fig.~\ref{fig:dp_extract_result_single} (c)) and found that this feature map was indeed highly activated by the white strip in the images of sheepdogs.
Our experts were impressed with this finding, because the datapath extraction algorithm not only finds a commonsensical feature map, but also finds unexpected feature map(s).
%, which is useful for understanding the .

% Although our datapath extraction approach is able to extract critical neurons for classifying an image, it sometimes lacks representativeness
\noindent \textbf{\normalsize Comparison between the extracted datapaths for one example and a set of examples.}
As shown in Fig.~\ref{fig:dp_extract_set}, we compared the extracted datapath for an image with only one panda face (Fig.~\ref{fig:dp_extract_set}A) with the one for a set of images that contain both panda faces and whole pandas (Fig.~\ref{fig:dp_extract_set}B).\looseness=-1

We found that the most critical feature maps extracted for an image with only one panda face contain the feature maps that detect a panda face (Fig.~\ref{fig:dp_extract_set}C and D) and important features on a panda face, such as eyes and ears (Fig.~\ref{fig:dp_extract_set}E).
The neuron in the blue rectangle (Fig.~\ref{fig:dp_extract_set}E) was unexpected at first glance.
After examining more discrepancy maps for that feature map and the activation heat map (Fig.~\ref{fig:dp_extract_set}F), we discovered that the feature map learned to detect black dots.
Such \doc{a} feature is a recognizable \doc{attribute of} a panda face.
%feature

\doc{Just like} the datapath of a panda's face, the datapath extracted for a set of panda images includes the feature maps for detecting a panda face (Fig.~\ref{fig:dp_extract_set}G and H).
 %Same as
In addition to these feature maps, the most critical feature maps include a feature map for detecting the black-and-white body of a panda (Fig.~\ref{fig:dp_extract_set}I), which is also an important feature for classifying a panda.
We visualized the activations on this feature map to verify that the feature map indeed learns to detect a panda body (Fig.~\ref{fig:dp_extract_set}J).\looseness=-1

This finding echoes requirement \textbf{R1} in Sec.~\ref{sec:require}, i.e., \doc{that} the extracted critical feature maps for one example may not be representative for a given set of images with the same class label.

\subsection{Case Study}
\label{sec:case}

There are two main types of adversarial attacks: white-box attack and a black-box attack~\cite{Akhtar2018_threat}.
The white-box attack means that the attacker has a full knowledge of the target model, including the parameters, architecture, training method, and even the training data.
While the Black-box attack assumes that the attacker knows nothing about the target model, \rev{which can be used to evaluate the transferability of adversarial examples}.
Next, we demonstrate how AEVis helps \rev{experts E$_1$ and E$_2$} analyze both white-box and black-box adversarial examples.
In both case studies, we utilized the dataset that is from the NIPS 2017 non-targeted adversarial attack and targeted adversarial attack competitions~\cite{NIPS2017_Challenge}, \rev{because the experts are familiar with this dataset.}  
It contains 1,000 images (299 pixel $\times$ 299 pixel)
%It contains 1,000 images, each of which is a 299 pixel $\times$ 299 pixel color image.
%For brevity’s sake, we refer to this dataset as the NIPS dataset.
We employed the white-box non-targeted attacking method developed by the winning machine learning group~\cite{NIPS_attack_code,dong2017boosting} to generate one adversarial example for each image in the dataset.
%The set of adversarial examples is denoted as the NIPS-ADV dataset.

% Transferability refers to the ability of an adversarial
% example to remain effective even for the models
% other than the one used to generate it.

% Black-box attacks feed a targeted model with the ad-
% versarial examples (during testing) that are gener-
% ated without the knowledge of that model. In some
% instances, it is assumed that the adversary has a
% limited knowledge of the model (e.g. its training
% procedure and/or its architecture) but definitely does
% not know about the model parameters. In other
% instances, using any information about the target
% model is referred to as ‘semi-black-box’ attack. We
% use the former convention in this article.

% the complete knowledge of
% the targeted model, including its parameter values,
% architecture, training method, and in some cases its
% training data as well.

\subsubsection{Analyzing White-Box Adversarial Examples}

\noindent \textbf{\normalsize Model.}
\shixia{One of the state-of-the-art CNNs for image classification, ResNet101~\cite{He2016_CVPR_residual}, is employed as the target model.}
It contains 101 layers. 
We used the pre-trained model from TensorFlow~\cite{Tensorboard2017_web}.
It achieves a high accuracy ($96.3\%$) on the normal examples, and a very low accuracy ($0.9\%$) on the generated adversarial examples.

\begin{figure}[htbp]
  \centering
  \vspace{-2mm}
  \includegraphics[width=\linewidth]{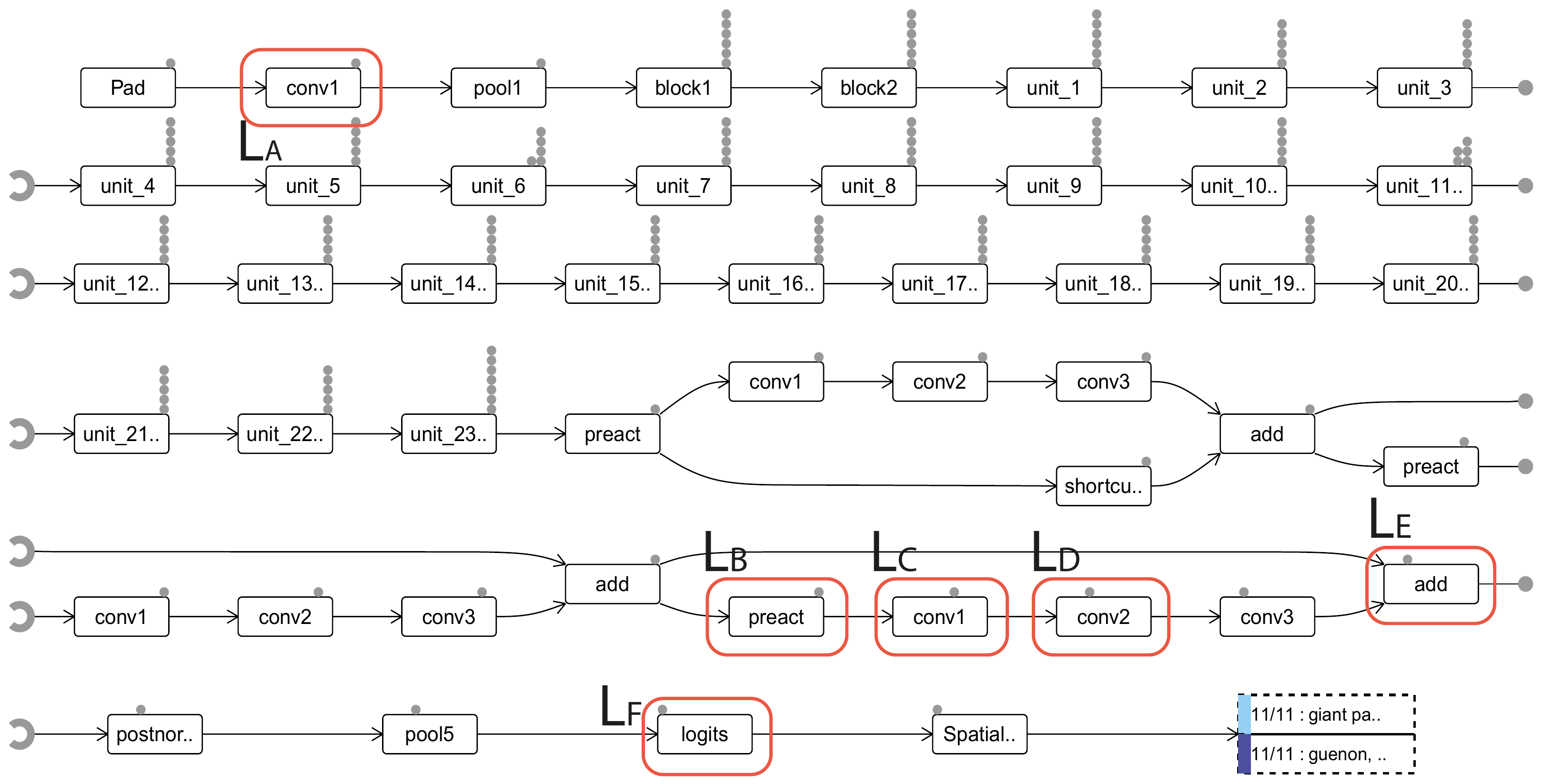}
    \vspace{-2mm}
  \caption{The datapath overview illustrating the activation differences between the datapaths of the adversarial examples and normal examples.\looseness=-1}
%     \vspace{-2mm}
  %(c) critical neurons found by our back-propagation based approach.}\looseness=-1
  \label{fig:case0_overview}
\end{figure}

% \begin{figure}[htbp]
%   \centering
% %   \vspace{-4mm}
%   \includegraphics[width=\linewidth]{fig/case0_illus.pdf}
%   \caption{A summary of the explanation for why the adversarial panda image was not classified as a panda.}
%   %(c) critical neurons found by our back-propagation based approach.}\looseness=-1
%   \label{fig:case0_illus}
% \end{figure}

\noindent \textbf{\normalsize Analysis process.}
To start the analysis, we calculated an adversarial score for each image~\cite{pang_towards}.
A high score means the image is most probably to be an adversarial example.
The expert \rev{E$_1$} then focused on the most uncertain images with medium adversarial scores.
% In particular, in the union of the NIPS and NIPS-ADV dataset, the top-100 images with the largest adversarial scores are all adversarial examples, and 94 images in the 100 images with lowest adversarial scores are normal images.
After examining these uncertain adversarial examples, \rev{E$_1$} selected one for further investigation ($I_B$ in Fig.~\ref{fig:teaser}).
It contains a panda head but is misclassified as a guenon monkey.
To understand the root cause of this misclassification, he wanted to compare its datapath with the one of the corresponding normal example ($I_A$ in Fig.~\ref{fig:teaser}).
To improve the representativeness of the extracted datapath for the normal example (Sec.~\ref{sec:eval_dp}, Fig.~\ref{fig:dp_extract_set}), he added 10 more normal panda images as well as the corresponding adversarial examples, to the images of interest (\textbf{R1}).
Each added adversarial example is misclassified as a guenon monkey.
Accordingly, \rev{E$_1$} split these images into two sets.
The first set (adversarial group) contained 11 adversarial examples.
% each was misclassified into a different class.
% Thus, there were 11 datapaths in this group.
The second set of images (normal group) contained 11 normal examples.
Then he extracted and visualized the datapaths for these two sets of images.\looseness=-1
% We calculated the common datapath for this group of images.
% As comparing 12 (11+1) datapaths brings too much burden to the expert, he selected two representative datapath from the adversarial group and compared them with the datapath extracted for the normal group.

\begin{figure}[!t]
  \centering
%   \vspace{-4mm}
  \includegraphics[width=0.9\linewidth]{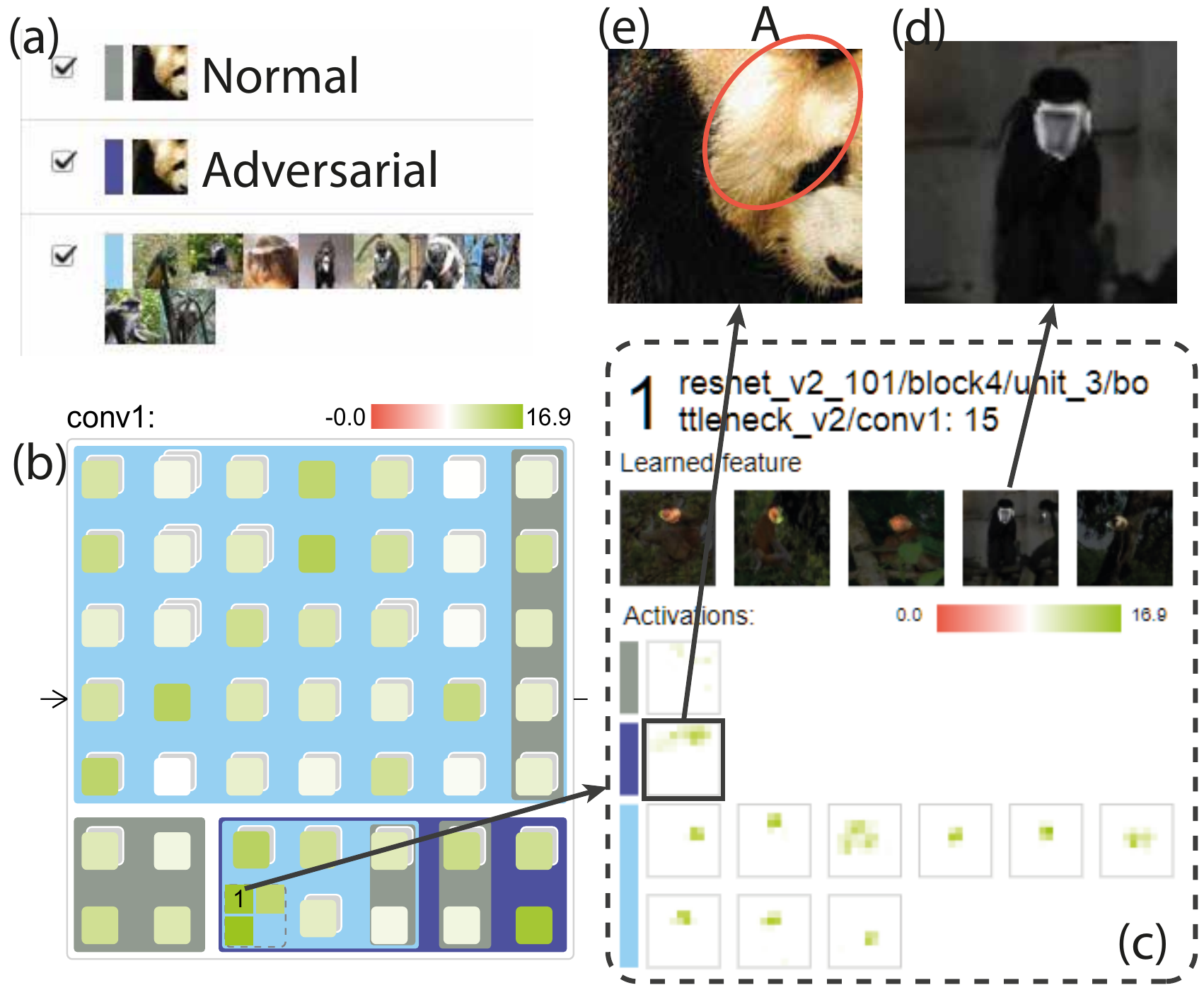}
  \vspace{-2mm}
  \caption{Explanation of why the adversarial panda image is misclassified as a monkey. The CNN erroneously detects a monkey's face: (a) input images; (b) feature maps; (c) neurons; (d) learned feature; and (e) adversarial example.}
  %(c) critical neurons found by our back-propagation based approach.}\looseness=-1
  \label{fig:case0_monkey}
  \vspace{-4mm}
\end{figure}

% The expert compared the extracted datapaths for these two sets of images.
The datapath overview is shown in Fig.~\ref{fig:case0_overview}.
The dot plots encoded the activation similarity between these two datapaths.
The expert \rev{E$_1$} found that at the bottom layer of the network (Fig.~\ref{fig:case0_overview}$L_A$), the activation similarity is almost 1.0 (the dot is on the right).
This similarity remained to be 1.0  until layer $L_B$ in Fig.~\ref{fig:case0_overview}.
After layer $L_B$, layer $L_C$ appear\rev{ed} as a ``diverging point'', where the activation similarity largely decreased (\textbf{R4}).
In the following layers, the activation similarity continued decreasing to 0 ($L_D$, $L_E$, $L_F$), which resulted in the misclassification.

To analyze which feature map is critical for the divergence, the expert \rev{E$_1$} expanded the diverging point $L_C$ (Fig.~\ref{fig:teaser}).
% By examining the size of each cell in the treemap (Fig.~\ref{fig:teasear}), he found that not only the activation similarity, but also the topological similarity among the datapaths largely decreased at the layer $L_C$.
% This verifies that the layer $L_C$ is a diverging point.
%\shixia{For Shixia: Not sure this is true for comparing 3 datapaths.}
In addition, he set the color coding of each feature map as the activation difference between two sets of examples (activation difference $=$ activations of normal images $-$ activations of adversarial examples).
A large activation difference indicates that the corresponding feature map detects its learned feature in the normal images but did not detect such a feature in the adversarial examples. 
Because the feature map $F_C$ in Fig.~\ref{fig:teaser} shows the largest activation difference, he then checked its neurons.
By examining the learned feature of the neurons (Fig.~\ref{fig:teaser}A), he discovered that the neurons learned to detect a black patch that resembles a panda's ear (Fig.~\ref{fig:teaser}B and C).
Such a feature is critical for detecting a panda's face, which does not appear in the adversarial example.
% By further examining the detailed neuron activations, he found that the feature map did not detect there  adversarial
To further investigate the influence of this feature map, \rev{E$_1$} continued to expand the next layer (layer $L_D$).
He found that there was a large activation difference on the feature map for detecting a panda's face ($F_D$ in Fig.~\ref{fig:teaser}).
This indicates that the corresponding feature map, $F_D$, fails to detect a panda's face, which is a direct influence of the large difference on $F_C$.
By the same analysis, \rev{E$_1$} found that in the layer that detected the highest-level features ($L_E$), there was also a large activation difference on the feature map for detecting a panda's face. 
As the target CNN cannot detect a panda's face in highest-level features, the CNN failed to classify the adversarial example as a panda image.

The above analysis explains why the CNN failed to classify the adversarial example as a panda, but cannot explain why the adversarial example is classified as a monkey.
Thus, \rev{E$_1$} compared the adversarial example with the corresponding example and a set of normal monkey images (Fig.~\ref{fig:case0_monkey} (a)).
\rev{Inspired} by the above analysis, \rev{E$_1$} directly expanded layer $L_C$ (diverging point), and examined its feature maps (Fig.~\ref{fig:case0_monkey} (b)).
% He was interested why these two sets of images resulted in the same prediction.
% Thus, he focused on the shared critical feature maps of these two datapaths.
% By examining the most important feature map for classifying a monkey in this layer (Fig.~\ref{fig:case0_monkey} (b)), the expert discovered that this feature map learned to detect the face of a monkey (Fig.~\ref{fig:case0_monkey} (c) and (e)).
After checking the activations of the adversarial example on the ``monkey's datapath'', the expert found that the activations were the largest on the feature map for detecting the face of a monkey (Fig.~\ref{fig:case0_monkey} (c) and (d)). 
% on this feature map is the largest among all the critical feature maps in the corresponding datapath.
This behavior of the CNN was unexpected because there was no monkey face in the adversarial example at first glance.
Thus, \rev{E$_1$} examined the detailed neuron activations on this feature map and discovered that the high activations appeared in the top right corner of the neurons, which corresponded to the top right corner of the image (Fig.~\ref{fig:case0_monkey} (e)).
\rev{E$_1$} did not understand why the CNN detected a monkey face at that part of the image.
By carefully examining the adversarial example and the learned features of the neurons, he finally figured out the reason: there is a dark strip with a bright strip on each side (Fig.~\ref{fig:case0_monkey}A).
It is a recognizable attribute for the face of a guenon monkey, which made the CNN mispredict the adversarial example as a monkey.

\begin{figure}[!t]
  \centering
%   \vspace{-4mm}
  \includegraphics[width=\linewidth]{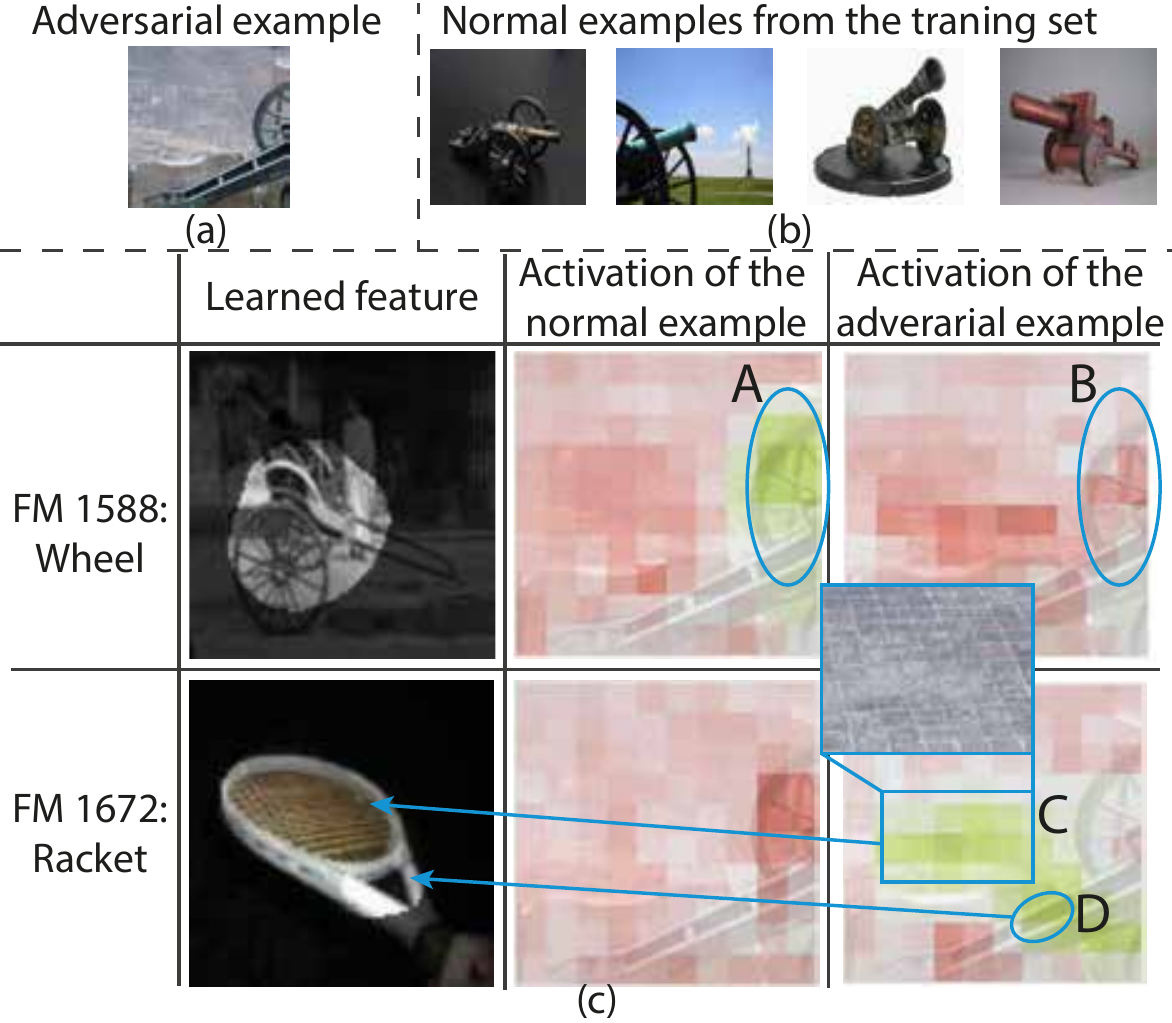}
  \vspace{-4mm}
  \caption{Explanation of why the adversarial cannon image is misclassified as a racket: (a) the adversarial example image; (b) normal examples from the training set; (c) the learned features and activations of neurons in the layer L$_E$ in Fig.~\ref{fig:case0_overview}.}
  %(c) critical neurons found by our back-propagation based approach.}\looseness=-1
  \label{fig:case0_cannon}
  \vspace{-4mm}
\end{figure}

\rev{Another case is shown in Fig.~\ref{fig:case0_cannon}, where a cannon image is misclassified as a racket (probability: 0.997).
The reasons behind the misclassification are two-fold.
First, the CNN recognizes a large wheel in the normal image (Fig.~\ref{fig:case0_cannon}A) but cannot recognize a large wheel in the adversarial image (Fig.~\ref{fig:case0_cannon}B), while a large wheel is a recognizable attribute of a cannon (Fig.~\ref{fig:case0_cannon} (b)).
Second, E$_1$ found that the CNN recognizes the head of a racket (Fig.~\ref{fig:case0_cannon}C),
which is connected to the throat of a racket (Fig.~\ref{fig:case0_cannon}D). 
These two reasons lead to the misclassification.}

\rev{E$_1$ commented, ``There may be different subtle causes that lead to the misclassification of different adversarial examples.
The value of AEVis is that it helps me find such causes quickly and effectively.
I can easily integrate my knowledge into the analysis process by leveraging the provided interactive visualizations.
I also see a great opportunity to use AEVis in analyzing more adversarial examples and summarize the major causes of the misclassification. 
This probably benefits future research on robust deep learning.''}

% We also discussed with the experts about the generalization of the findings in the case studies.
% Expert E$_1$ commented that AEVis is able to disclose the root cause of the misclassification produced by a set of images.
% But he is not sure whether the found reasons can generalize to other unexamined images or other adversarial example generation approaches.
% In these scenarios, other reasons could lead to the misclassification and he would like to try to categorize the major causes of the misclassification with the help of AEVis. 
% }

% \mengchen{TODO: image substraction}
% \mengchen{The attacking algorithm detects a part of the image and }

% Although the CNN detects a monkey's face in the adversarial example, if it still have high activations XXXX. it will be classified as a panda.
% Thus, the expert further compared the datapaths of the adversarial example and the corresponding normal panda image ().

% \begin{figure}[htbp]
%   \centering
% %   \vspace{-4mm}
%   \includegraphics[width=\linewidth]{fig/case2_overview.pdf}
%   \caption{The overview of comparing the datapaths for the adversarial panda image and beagle images.}
%   %(c) critical neurons found by our back-propagation based approach.}\looseness=-1
%   \label{fig:case2_overview}
% \end{figure}

\subsubsection{Analyzing Black-box Adversarial Examples}

% \noindent \textbf{\normalsize Data.}
% Same as the data in the above case study.

\noindent \textbf{\normalsize Model.}
We used the VGG-16~\cite{simonyan2014very} network to analyze black-box attacks because our employed attacking method has the knowledge of a set of state-of-the-art CNNs, such as ResNet~\cite{He2016_CVPR_residual} and the Inception network~\cite{szegedy2016rethinking}.
Thus, we adopted a traditional CNN (VGG) to analyze black-box attacks. 
We utilized the pre-trained VGG-16 network from TensorFlow~\cite{Tensorboard2017_web}.
It achieved a $84.8\%$ accuracy on the normal examples, and a $22.4\%$ accuracy on the adversarial examples.
% Image preprocessing follows the approach recommended in TensorFlow.

\noindent \textbf{\normalsize Analysis process.}
The expert \rev{E$_2$} continued to explore the analyzed adversarial example, and tested it on the VGG network.
He found that it was misclassified as a beagle (a type of dog, Fig.~\ref{fig:case2_dog} (a)).
%It indicated that the black-box attack is successful for this image.
To check how the prediction was made, \rev{E$_2$} extracted and compared the datapaths for normal beagle images and  the adversarial panda image.
He found that the there was no obvious ``merging point'' in VGG-16, because these images do not truly ``merge'' due to the high prediction score of beagle images (0.75$-$1.0) and the \shixia{relatively} low prediction score of the adversarial example (0.46). %because the activation similarity gradually decreased to 0 from the middle layers.
Thus, \rev{E$_2$} relied on his knowledge to select the layer that detected the highest-level features (Fig.~\ref{fig:case2_dog} (b)).
By setting the color coding of feature maps as the activation of the adversarial example, \rev{E$_2$} found a large activation appeared on a feature map in shared feature maps between the beagle's and adversarial panda's datapath (Fig.~\ref{fig:case2_dog}A).
It is a potential cause that leads to the misclassification.
Thus, he examined the learned feature of the neurons in this feature map (Fig.~\ref{fig:case2_dog} (c)), and found that they learned to detect the black nose of a beagle (a black patch, Fig.~\ref{fig:case2_dog}B).
This feature is a recognizable attribute for a beagle.
To understand why the adversarial example caused a large activation on this feature map, he further examined the activation heat map, and found that the neurons in the top right corner are highly activated (Fig.~\ref{fig:case2_dog}C).
It indicates that there is such a feature in the corresponding part of the image.
In particular, that part of the images is the black ear of the panda, which is also a black patch (Fig.~\ref{fig:case2_dog}C).
This feature misled the VGG-16 network to detect a black nose in the adversarial panda image, which then led to the misclassification.\looseness=-1

\begin{figure}[htbp]
  \centering
%   \vspace{-2mm}
  \includegraphics[width=\linewidth]{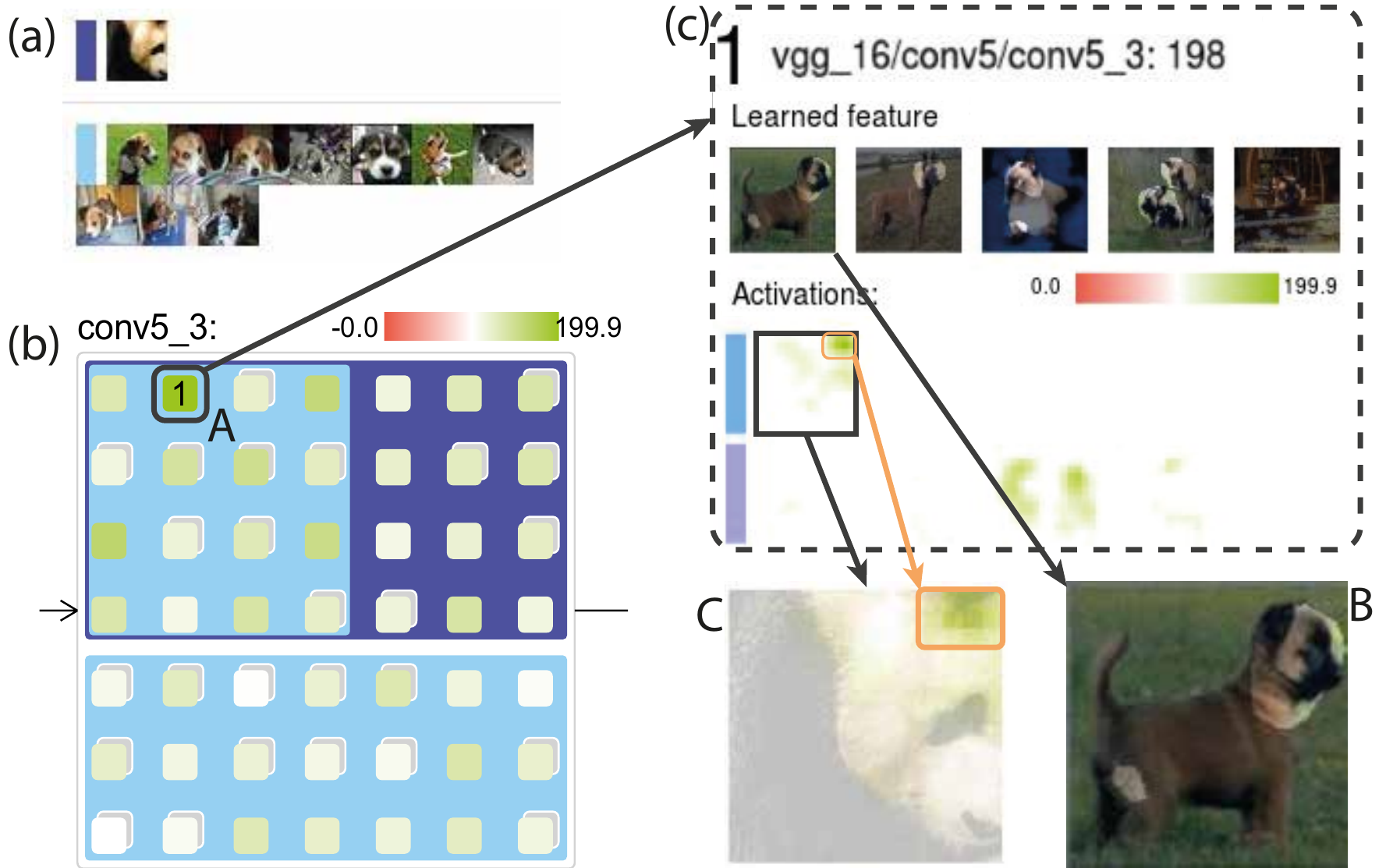}
  \vspace{-4mm}
  \caption{The explanation of why the adversarial panda image was misclassified as a beagle (dog) by the VGG-16 network: (a) input images; (b) feature maps; and (c) neurons.}
  \vspace{-4mm}
  %(c) critical neurons found by our back-propagation based approach.}\looseness=-1
  \label{fig:case2_dog}
\end{figure}

\section{Discussion}

AEVis can better disclose the inner workings of adversarial examples and help discover the vulnerable neurons \doc{that lead} to incorrect predictions. 
%leading
However, it also comes with several limitations, which may shed light on future research \doc{directions}.
%In this section, we discuss scalability and generalization of AEVis. 

% \subsection{Scalability}

\noindent \textbf{\normalsize Scalability.}
We have demonstrated that AEVis is able to analyze a state-of-the-art CNN (ResNet101), which has 101 layers and is much deeper than traditional CNNs (e.g., VGG-Net).
More recently, researchers have developed many deeper CNNs with thousands of layers~\cite{He2016_CVPR_residual}.
When handling such deep DNNs, if an expert zooms in to low levels of the layer hierarchy, the layers of interest cannot fit in one screen, even with the help of our segmented DAG.
To alleviate this issue, we can employ a mini-map to help the expert track \doc{the} current viewpoint, which has been proven effective in TensorFlow Graph Visualizer~\cite{wongsuphasawat2018visualizing}.
The dot plot is another factor that hinders AEVis from analyzing CNNs with thousands of layers.
\doc{This is because} a layer group may contain hundreds of layers, and the height of the \doc{resulting} dot plot will be too large, which is a waste of screen space.
 %The reason is that resulted
To solve this problem, we can use non-linear dot plots~\cite{rodrigues2018nonlinear} to improve the scalability of AEVis.\looseness=-1

Currently, we utilize a Euler-diagram-based design to illustrate the overlapping relationship between/among datapaths.
Such \doc{a} design is suitable for comparing several datapaths~\cite{alsallakh2014visualizing}.
Although researchers found that approximately four objects can
be tracked in \doc{a} visual comparison~\cite{intriligator2001spatial,wang2016topicpanorama,yantis1992multielement}, experts may have special needs of comparing a lot of datapaths.
To fulfill these needs, we can leverage more scalable set visualization techniques, such as PowerSet~\cite{alsallakh2017powerset}.
%In addition, even with the help of , all the layers cannot 
%To solve this problem, we can leverage 

% Currently, simply adding more layers to a deep model often leads to performance degradation due to over-fitting~\cite{He2016_CVPR_residual}.
% When researchers successfully solve this problem and design a deep model with thousands of layers
% To handle deep the visualization  

\noindent \textbf{\normalsize Generalization.}
AEVis aims at analyzing the adversarial examples of CNNs because most research on adversarial attacks focuses on generating adversarial images for CNNs.

In addition to attacking CNNs, there are several initial attempts to attack other types of DNNs~\cite{Akhtar2018_threat}, such as multilayer perceptron (MLP), recurrent neural networks (RNNs), autoencoders (AEs), and deep generative models (DGMs).
\rev{Among these models, AEVis can be directly used to analyze MLPs by treating each neuron as a feature map that contains one neuron.
For other types of DNNs, we need to develop suitable datapath extraction and visualization methods.}
%Ori: In these types of DNNs, there are also datapaths.
For example, Ming et al.~\cite{ming2017understanding} demonstrated that some neurons in an RNN were critical for predicting the sentiment of a sentence, such as the neurons for detecting positive/negative words.
Such neurons and their connections form a datapath for an RNN.
Thus, by extracting and visualizing datapaths, AEVis can be extended to analyze the root cause of adversarial examples for these types of DNNs.
%Ori: To this end, we need to develop suitable datapath extraction and visualization approaches for different types of DNNs.
% Based on the extracted datapaths, we need to develop suitable visualizations for different types of DNNs.  
For example, to visualize the datapath of RNNs, we can first unfold the architecture of an RNN to a DAG~\cite{goodfellow2016deep}, and then employ a DAG layout algorithm to calculate the position of each unfolded layer.

In addition to images, researchers try to generate adversarial examples for other types of data~\cite{Akhtar2018_threat}, such as adversarial documents and adversarial videos.
To generalize AEVis to other types of data, we need to change the visual hint for neurons (discrepancy map and activation heat map) according to the target data type.
% the datapath view (Fig.~\ref{fig:teaser}) needs not to be changed.
% The major change will in 
For example, \doc{when} analyzing adversarial documents, we can use a word cloud to represent the ``learned feature'' of a neuron in an RNN~\cite{ming2017understanding}.
%for
In the word cloud, we select the keywords that activate the neuron.

\section{Conclusion}
We have presented a robustness-motivated visual analytics approach that helps experts understand the inner workings of adversarial examples and diagnoses the root cause of incorrect predictions introduced by the adversarial examples. 
The major feature of this approach is that it centers on the concept of datapaths to tightly combine datapath extraction and visualization. 
Two case studies were conducted with two machine learning experts to demonstrate the effectiveness and usefulness of our approach in analyzing both white-box and black-box adversarial examples. 

One interesting avenue for future \doc{research} is to monitor the online testing process, detect potentially adversarial examples, and remove  them from any further processing. 
%work
The key is to design a set of streaming visualizations that can incrementally integrate the \doc{incoming} log data with \doc{existing data}. 
%the existing one new coming
We would also like to continue working with the machine learning experts to conduct several field experiments, \doc{with the aim of} designing  more targeted and effective defense solutions based on the discovered root cause. 
%which aims at 
Another important direction is to analyze the robustness of other types of DNNs, such as RNNs and DGMs. 
For these types of DNNs, exciting research topics include more efficient datapath extraction algorithms and suitable visualizations  for different types of DNNs. 
}

%% if specified like this the section will be ommitted in review mode
\acknowledgments{
% Yinpeng
This research was funded by National Key R\&D Program of China (No. SQ2018YFB100002), the National Natural Science Foundation of China (No.s 61761136020, 61672308, 61620106010, 61621136008, 61332007), Beijing NSF Project (No. L172037), Microsoft Research Asia, Tiangong Institute for Intelligent Computing, NVIDIA NVAIL Program, and Tsinghua-Intel Joint Research Institute.
The authors would like to thank Yinpeng Dong for insightful discussions in the case studies and Jie Lu for help in the development of AEVis.
\looseness=-1}
%\newpage
%\clearpage
\small
\bibliographystyle{abbrv}
%\addtolength{\itemsep}{5ex}
\bibliography{reference}

% sigproc.bib is the name of the Bibliography in this case

%1 page

\end{document}